\documentclass[pdflatex]{sn-jnl}

\usepackage[authoryear]{natbib}
\usepackage{multirow}
\usepackage{amsmath,amssymb}
\usepackage{xcolor}
\usepackage{booktabs}
\usepackage{algorithm}
\usepackage{algpseudocode}
\newcommand{\mytitle}{Nonmyopic Global Optimisation via Approximate Dynamic Programming}
\hypersetup{
    pdfauthor=Filippo Airaldi,
    pdfcreator=Filippo Airaldi,
    pdftitle=\mytitle,
}
\usepackage{cleveref}
\crefformat{figure}{Fig.~#2#1#3}
\Crefformat{figure}{Fig.~#2#1#3}
\usepackage[inline]{enumitem}
\usepackage{tikz}
\usetikzlibrary{calc,positioning,shapes.geometric}
\usepackage{pgfplots}
\usepgfplotslibrary{fillbetween,groupplots}
\usepackage{pgfplotstable}
\pgfplotsset{compat=1.18}
\def\axisdefaultwidth{\textwidth}  
\usepackage{orcidlink}
\usepackage[short,nocomma]{optidef}
\usepackage{siunitx}
\usepackage[normalem]{ulem}
\usepackage{pdflscape,longtable,afterpage}  

\DeclarePairedDelimiter{\norm}{\lVert}{\rVert}
\DeclarePairedDelimiter{\stochasticargument}{\{}{\}}
\NewDocumentCommand{\expval}{e{_}}{%
    \mathbb{E}\IfValueT{#1}{_{#1}}\stochasticargument
}
\newtheorem{remark}{Remark}

\definecolor{C0}{RGB}{90,91,159}  
\definecolor{C1}{RGB}{255,111,97}  
\definecolor{C2}{RGB}{217,79,112}  
\definecolor{C3}{RGB}{0,148,115}  
\definecolor{C4}{RGB}{128,128,128}  
\definecolor{C5}{RGB}{240,192,90}  
\definecolor{darkslategray}{RGB}{38,38,38}
\definecolor{lavender}{RGB}{234,234,242}
\pgfplotsset{
    every axis/.style={
        axis background/.style={fill=lavender},
        axis line style={white},
        tick align=outside,
        tick pos=left,
        enlarge x limits=0.05,
        enlarge y limits=0.05,
        tick scale binop=\times,
        x grid style={white},
        y grid style={white},
        xmajorgrids,
        ymajorgrids,
        xtick style={color=darkslategray},
        ytick style={color=darkslategray},
    },
    every axis title/.style={anchor=south, at={(0.5,1)}},
    every axis legend/.append style={
        cells={anchor=west},
        fill=lavender,
        draw=white,
    },
    error bars/.cd,
    x dir=both,
    x explicit,
    y dir=both,
    y explicit,
}

\title[\mytitle]{\mytitle}
\author*{
    \fnm{Filippo} \sur{Airaldi}\orcidlink{0000-0001-6595-2932}
}\email{f.airaldi@tudelft.nl}
\author{
    \fnm{Bart} \sur{De Schutter}\orcidlink{0000-0001-9867-6196}
}\email{b.deschutter@tudelft.nl}
\author{
    \fnm{Azita} \sur{Dabiri}\orcidlink{0000-0002-4051-4896}
}\email{a.dabiri@tudelft.nl}
\affil{\orgdiv{Delft Center for Systems and Control}, \orgname{Delft University of Technology}, \orgaddress{\street{Mekelweg 2}, \postcode{2628 CD} \city{Delft}, \country{The Netherlands}}}

\abstract{
    Global optimisation to optimise expensive-to-evaluate black-box functions without gradient information. Bayesian optimisation, one of the most well-known techniques, typically employs Gaussian processes as surrogate models, leveraging their probabilistic nature to balance exploration and exploitation. However, these processes become computationally prohibitive in high-dimensional spaces. Recent alternatives, based on inverse distance weighting (IDW) and radial basis functions (RBFs), offer competitive, computationally lighter solutions. Despite their efficiency, both traditional global and Bayesian optimisation strategies suffer from the myopic nature of their acquisition functions, which focus solely on immediate improvement neglecting future implications of the sequential decision making process. Nonmyopic acquisition functions devised for the Bayesian setting have shown promise in improving long-term performance. Yet, their combination with deterministic surrogate models remains unexplored. In this work, we introduce novel nonmyopic acquisition strategies tailored to IDW and RBF based on approximate dynamic programming paradigms, including rollout and multi-step scenario-based optimisation schemes, to enable lookahead acquisition. These methods optimise a sequence of query points over a horizon (instead of only at the next step) by predicting the evolution of the surrogate model, inherently managing the exploration-exploitation trade-off via optimisation techniques. The proposed approach represents a significant advance in extending nonmyopic acquisition principles, previously confined to Bayesian optimisation, to deterministic models. Empirical results on synthetic and hyperparameter tuning benchmark problems, a constrained problem, as well as on a data-driven predictive control application, demonstrate that these nonmyopic methods outperform conventional myopic approaches, leading to faster and more robust convergence.
}


\begin{document}

\maketitle

\section{Introduction} \label{section:introduction}

Global optimisation (GO) is a class of optimisation techniques designed for finding the global minimum (or maximum) of an objective function. hese techniques are especially relevant when the objective is, e.g., a black-box function whose gradient information cannot be computed, or is highly irregular and non-convex \citep{horst_handbook_1995,rios_derivative_2013,joaquim_engineering_2022,audet_derivative_2025}. These aspects make GO strategies appealing and useful in domains where derivative-based methods struggle or fail, e.g., when the objective function is non-smooth or its gradient too costly to evaluate. Examples of applications include hyperparameter selection in machine learning algorithms \citep{snoek_practical_2012}, automatic tuning of closed-loop controllers \citep{sorourifar_data_2021}, and optimisation of neural functions \citep{strong_global_2023}. Well-known methodologies in this field include particle swarm optimisation \citep{kennedy_particle_1995,zhan_adaptive_2009}, genetic algorithms \citep{Whitley_genetic_1994} and evolutionary algorithms \citep{hansen_completely_2001}.

Despite their applicability to a wide range of problems, when dealing with prohibitively slow- and/or expensive-to-evaluate functions, the aforementioned methods are often not suitable as they may need to inquire the function value at a large numbers of different query points. In these cases, it is in general preferred to leverage surrogate-based GO techniques, whose primary goal is to minimise the number of function evaluations while ensuring that the algorithm escapes local optima and converges to a global solution or, at least, to a near-optimal minimum \citep{joaquim_engineering_2022,audet_derivative_2025}. To this end, surrogate-based optimisation comprises two main ingredients: a surrogate model that approximates the underlying unknown objective function and an acquisition function, usually in the form of a cheaper-to-solve optimisation subproblem, that guides the sampling of new points to be observed and governs the fundamental trade-off between exploration (searching for better solutions in unexplored regions) and exploitation (focusing on convergence in the current most promising region). An iterative algorithm integrates these two components by first updating the surrogate model with the latest information and then solving the acquisition for the next query point, and it runs until a stopping criterion is met, e.g., a maximum number of iterations or satisfactory convergence to a minimum. 

One of the most prominent techniques in surrogate-based GO is Bayesian optimisation (BO), which takes a probabilistic approach to the problem \citep{garnett_bayesoptbook_2023}. In BO, evaluations of the unknown function at any query point are assumed to be random variables, and the uncertainty of the objective over the unexplored input space is modelled as a probability distribution. The key to BO is the construction of a prior belief in the form of a probabilistic model, typically based on Gaussian processes (GPs) \citep{rasmussen_gaussian_2005}, which provides a stochastic estimate of the function at any point in the input space. As the noisy objective evaluations are iteratively collected, the prior is updated into a posterior which, in turn, is used to construct the acquisition function whose solution determines the next query point. While BO has been widely successful, the reliance on GP models comes with a computational burden, particularly in high-dimensional spaces.

Recently, alternative optimisation strategies that employ inverse distance weighting (IDW) \citep{joseph_regression_2011} and radial basis functions (RBFs) \citep{mcdonald_global_2007} have emerged as competitive approaches. Compared to BO, surrogate models based on IDW and RBF functions offer deterministic approximations of the black-box function and of the uncertainty between the surrogate and the unknown objective. Since no posterior computation is needed, an immediate benefit of this approach is that the resulting algorithms are lightweight and computationally efficient while still retaining a performance that is comparable to BO \citep{joseph_regression_2011,costa_rbf_2018,bemporad_global_2020}. Previous work in \cite{gutmann_radial_2001,regis_constrained_2005} has laid the theoretical foundations and shown their efficacy; more recently, their use has been successfully extended to other domains, such as preference learning \citep{bemporad_global_2021} and active learning \citep{bemporad_active_2023}.

However, despite the success of these traditional GO and BO strategies, much of the research has highlighted a crucial weak point: the myopic nature of its acquisition functions. Most standard acquisition functions, such as expected improvement (EI) by \cite{jones_efficient_1998} and upper confidence bound in  \cite{srinivas_gaussian_2010}, make decisions based solely on the current state of the surrogate model without considering the long-term implications of future query-evaluation data. Even a more sophisticated stratagem like the knowledge-gradient policy suffers from this drawback, as it considers the expected increment in the value of information for only one step ahead \citep{frazier_knowledge_2008}. Such short-sighted acquisition functions are likely to lead to suboptimal solutions, especially in problems where the exploration-exploitation balance is critical for good convergence or the number of evaluations is limited. 

Addressing the myopia of acquisition functions is thus a parallel and vibrant thread of the current research in the field. In the context of BO, nonmyopic acquisition functions have been proposed that leverage concepts from the field of dynamic programming (DP) \citep{bertsekas_dynamic_2017}. In particular, \cite{lam_bayesian_2016} was among the first to propose the use of rollout, a well-known approximate DP algorithm \citep{bertsekas_rollout_2020,bertsekas_multiagent_2021}, to compute a querying strategy that can predict across a horizon of multiple iterations the evolution of the surrogate model as new evaluations of the unknown objective are observed. Briefly, it employs a base acquisition function with no look-ahead intrinsic property (e.g., EI) as reward, and then uses a rollout scheme to find the sequence of query points that maximises the accumulation of these rewards along the prediction horizon. Whereas the base function is greedy, i.e., it selects the next query point to maximise the immediate reward without regard to the remaining successive iteration, the resulting rollout strategy is able to take into consideration how hypothetical future information affects the sequential decision making process, thus acting nonmyopically. This ultimately results in \textit{multi-step lookahead} strategies that are often more performing and converge faster, and that can inherently balance the exploration-exploitation trade-off over the horizon in a systematic and structured way via optimisation techniques (though the extent of these benefits can vary from one application to another). 

More recent literature has sprouted from the work of \cite{lam_bayesian_2016}. Advanced nonmyopic strategies for BO have been devised that, e.g., improve the convergence of statistical estimates \citep{lee_efficient_2020}, adjust the length of the prediction horizon automatically \citep{yue_why_2020}, or tackle cost-constrained optimisation problems \citep{lee_nonmyopic_2021}. The formal problem statement and theory behind these methodologies have also been enhanced by \cite{bertsekas_rollout_2022}. Among the relevant nonmyopic approaches, noteworthy is GLASSES \citep{gonzalez_glasses_2016}, which builds an approximation of future expected rewards from a batch policy and uses it to mitigate greedy acquisitions; however, this policy is only a heuristic and not the optimal one. Likewise, BINOCULARS \citep{jiang_binoculars_2020} also leverages an approximation of future rewards that is then maximised, but this approximation is further shown to act as lower bound to the actual expected future realisations. \cite{wu_practical_2019} instead propose a two-step lookahead variant of EI, and show how the corresponding acquisition problem can be solved via gradient-based techniques. Finally, \cite{jiang_efficient_2020} provides a general framework for multi-step BO which enjoys an efficient and scalable implementation on GPUs. However, while it is apparent that this line of research is gaining momentum in the Bayesian community, its application to GO techniques with deterministic surrogate models, especially the ones based on IDW and RBF that reduce computation costs while maintaining comparable performance, remains relatively unexplored.

In this work, we propose to address this gap in the current literature by leveraging the successes of IDW- and RBF-based optimisation strategies and of nonmyopic BO strategies. We propose to employ DP methodologies to craft multi-step lookahead acquisition problems for GO strategies using deterministic IDW and RBF surrogate models. Specifically:
\begin{enumerate}
    \item We adapt the current IDW- and RBF-based GO optimisation procedure to the DP setting. This includes formulating the dynamics of these surrogate models, which describe how these approximation models evolve when new data is made available, and devising a proper reward signal. Furthermore, since (unlike GPs) these surrogate models do not generate a posterior, we propose a heuristic posterior that allows for sampling in the DP algorithms.  
    \item We propose two well-known algorithms, rollout and multi-step scenario-based optimisation, to solve our nonmyopic acquisition problems based on IDW and RBF via different sampling strategies.
    \item We validate the proposed nonmyopic schemes on several numerical benchmark cases of various dimensionality and difficulty, including synthetic and real-world hyperparameter tuning problems, as well as on a data-driven tuning experiment of a predictive controller based on closed-loop performance. The simulation results empirically confirm that the proposed methods can outperform the vanilla myopic GO strategies.
\end{enumerate}

The rest of the paper is organised as follows. \Cref{section:background} provides background information on myopic GO strategies and DP algorithms. The proposed nonmyopic methods are then presented and explained in \Cref{section:nonmyopic-go}. \Cref{section:numerical-experiments} provides two numerical experiments to validate these algorithms. Finally, \Cref{section:conclusions} concludes the
paper with some final remarks.

\section{Background} \label{section:background}

In this section, the foundations of the two main components of this work, i.e., global optimisation and dynamic programming, are presented.

\noindent \textbf{Notation.} In the context of nonmyopic/predictive algorithms, given the information available at time step (or iteration) $k$, the prediction of any quantity $y$ at time step (or iteration) $i \ge k$ is denoted as $\hat{y}_{i|k}$.

\subsection{Global Optimisation}  \label{section:background:globopt}

Given the function $f : \mathbb{R}^n \rightarrow \mathbb{R}$, we consider the following optimisation problem
\begin{argmini}
    { \scriptstyle x \in \mathcal{X} }{ f(x) }{ \label{eq:background:go:problem} }{ x^\star \in },
\end{argmini}
where $x$ is an $n$-dimensional vector over the compact set $\mathcal{X} \subseteq \mathbb{R}^n$. In particular, we assume that $f(x)$ is expensive to evaluate and no gradient information is available, while $\mathcal{X}$ encodes an arbitrarily constrained but known space. The goal is to estimate a solution $x^\star$ of \eqref{eq:background:go:problem} using an optimal sequential selection of observations based on feedback from preceding observations. As mentioned in the \nameref{section:introduction}, in GO this problem is tackled by modelling the objective function $f$ with a surrogate model. In line with \cite{bemporad_global_2020}, this work focuses on IDW \citep{joseph_regression_2011} and RBF \citep{mcdonald_global_2007} functions as surrogates of the black-box objective. To construct these surrogate models of the unknown function $f$, which will be referred to as $\hat{f}_k$, let us consider iteration $k \in \{N_0,\dots,N\}$ of the sequential estimation, with $N$ the maximum number of iterations and $N_0 \in [1, N)$ the number of initial warm-up iterations.\footnote{These $N_0$ iterations are often used to prime the regressor on an initial dataset of $N_0$ data points before starting the optimisation algorithm. This dataset can include a priori knowledge on $f(x)$, but more often is generated via randomly sampled evaluations of the objective. For the sake of simplicity, assume here $N_0=1$.} Assume $F_k = \begin{bmatrix} f_1 & \dots & f_k\end{bmatrix}^\top \in \mathbb{R}^k$ gathers the noiseless measurements $f_i = f(x_i)$ collected so far at the corresponding points $X_k = \begin{bmatrix} x_1 & \dots & x_k\end{bmatrix}^\top \in \mathbb{R}^{k \times n}$, with $x_i \neq x_j$, $\forall i,j \in \left\{1,\dots,k\right\}$, $i \neq j$. These query-observation pairs constitute the dataset $\mathcal{D}_k = \left\{ \left( x_i, f_i \right) \mid i = 1,\dots,k \right\}$ at the current iteration $k$. Naturally, these can also be defined recursively:
\begin{equation} \label{eq:background:go:recursive-dataset}
    \mathcal{D}_0 = \emptyset, \quad \mathcal{D}_{k+1} = \mathcal{D}_k \cup \left\{\left( x_{k+1}, f_{k+1}\right)\right\}.
\end{equation}

Given a generic query point $x \in \mathbb{R}^n$, IDW interpolation leverages the observed data to formulate, at current iteration $k$, the non-parametric approximation $\hat{f}_k : \mathbb{R}^n \rightarrow \mathbb{R}$
\begin{equation} \label{eq:background:go:idw-regression}
    \hat{f}_k(x) = \sum_{i=1}^k{ f_i v_{k,i}(x) } = F_k^\top V_k(x),
\end{equation}
where
\begin{align}
     v_{k,i}(x) &{}= \frac{ w_i(x) }{ \sum_{j=1}^k{ w_j(x) } }, & i=1,\dots,k, \\
     w_i(x) &{}= \frac{1}{\max{\left\{ d^2(x, x_i), \delta \right\}}}, & i=1,\dots,k, \label{eq:background:go:idw-weights}
\end{align}
and
\begin{equation}
    V_k(x) = \begin{bmatrix} v_{k,1}(x) & \dots & v_{k,k}(x) \end{bmatrix}^\top \in \mathbb{R}^k.
\end{equation}
We use $v_{k,i}$ to refer to the $i$-th element of $V_k$ at iteration $k$. In \eqref{eq:background:go:idw-weights}, we indicate with $d^2 : \mathbb{R}^n \times \mathbb{R}^n \rightarrow \mathbb{R}_{\ge 0}$ the squared Euclidean distance $d^2(x, x_i) = \norm{x - x_i}_2^2$ and $\delta > 0$ is a small scalar value that avoids division by excessively small numbers when $x \rightarrow x_i$. By inspection of \eqref{eq:background:go:idw-regression}, $v_{k,i}(x)$ can be interpreted as weighting the contribution of each observation $f_i$ based on the distance of the query point $x$ from $x_i$, hence the name IDW. Such formulation enjoys some favourable properties \citep[see][Lemma~1]{bemporad_global_2020}, but other choices of $w_i$ are available, e.g., $w_i(x) = \frac{e^{-d^2(x,x_i)}}{d^2(x,x_i)}$ is proposed in \cite{joseph_regression_2011} to intensify the rate of decay of each contribution with increasing distances.

Likewise, RBFs can be used to build a parametric surrogate function as
\begin{equation} \label{eq:background:go:rbf-regression}
    \hat{f}_k(x) = \sum_{i=1}^k{
        \beta_{k,i}\phi\bigl(\varepsilon d(x, x_i)\bigr)
    } = B_k^\top \Phi_k(x),
\end{equation}
with
\begin{align}
    B_k &{}= \begin{bmatrix} \beta_{k,1} & \dots & \beta_{k,k} \end{bmatrix}^\top \in \mathbb{R}^k, \\
    \Phi_k(x) &{}= \begin{bmatrix}
        \phi\bigl(\varepsilon d(x, x_1)\bigr) &
        \dots                                 &
        \phi\bigl(\varepsilon d(x, x_k)\bigr)
   \end{bmatrix}^\top \in \mathbb{R}^k,
\end{align}
where $\varepsilon > 0$ is a scalar parameter, $\phi : \mathbb{R} \rightarrow \mathbb{R}$ is an RBF, and $\beta_{k,i}$ is the $i$-th coefficient to be determined by solving the linear system
\begin{equation}
    F_k = M_k B_k,  \label{eq:background:go:rbf-linear-system}
\end{equation}
with the symmetric interpolation matrix
\begin{equation}
    M_k = \begin{bmatrix} \Phi_k(x_1) & \dots & \Phi_k(x_k) \end{bmatrix}  \in \mathbb{R}^{k \times k}.
\end{equation}
For the sake of conciseness, here we assume $M_k$ to be non-singular, so that a solution to \eqref{eq:background:go:rbf-linear-system} always exists. Note that methods to circumvent this issue are available, e.g., the singular value decomposition is commonly employed to invert $M_k$ while discarding singular values below a small threshold \citep{bemporad_global_2020}. RBFs have widely been used in function approximation schemes, e.g., in neural networks \citep{lowe_multivariable_1988}, since $\Phi_k$ offers a suitable basis for the function space from $\mathbb{R}^n$ to $\mathbb{R}$. In particular, in this work, the inverse quadratic basis function $\phi(y) = \frac{1}{1 + y^2}$ is employed (see \citet[Section~3.2]{bemporad_global_2020} for a collection of other popular choices for $\phi$).

The ensuing GO algorithm is carried out as follows. It starts at $k=N_0$ with the initial dataset $\mathcal{D}_{N_0}$ of query and observation pairs. At each iteration $k$, regression is performed on the current dataset $\mathcal{D}_k$ to obtain a surrogate model $\hat{f}_k$, which allows for the prediction of the unknown objective $f$ at any query point $x \in \mathcal{X}$. Next, the algorithm selects a new target point $x_{k+1}$ for evaluation, leading to observation $f_{k+1}$. The new pair $\left( x_{k+1}, f_{k+1} \right)$ is then added to the dataset $\mathcal{D}_k$, and a new iteration begins. The algorithm continues till the maximum number of iterations $N$ is reached (or another termination condition is met), at which point the approximate solution $x^\star \approx \max_{i=1,\dots,N+1} x_i$ to the original problem \eqref{eq:background:go:problem} is returned.

The selection of the next query point is encoded by the auxiliary problem
\begin{argmini}
    { \scriptstyle x \in \mathcal{X} }{ \Lambda\left(\mathcal{D}_k, x\right), }{ \label{eq:background:go:acquistion-problem} }{ x_{k+1} \in }
\end{argmini}
where $\Lambda : \left( \mathbb{R}^n \times \mathbb{R} \right)^k \times \mathbb{R}^n \rightarrow \mathbb{R}$ is often referred to as the \textit{acquisition function}. Compared to \eqref{eq:background:go:problem}, problem \eqref{eq:background:go:acquistion-problem} is assumed to be much cheaper and faster to solve, and does not involve evaluating the expensive objective function $f$ but rather only depends on the current dataset $\mathcal{D}_k$ and surrogate model $\hat{f}_k$. Thus, the acquisition function is assigned to the paramount task of guiding the optimisation towards an approximate global optimum while balancing the trade-off between exploration and exploitation.

Commonly, acquisition functions in BO exploit the posterior covariance associated with the GP approximator to take exploration into account naturally. However, since this work focuses on deterministic predictors, we borrow the acquisition function suggested and described in \cite{joseph_regression_2011,bemporad_global_2020} that is crafted ad hoc for IDWs and RBFs to induce the exploration-exploitation trade-off. For this purpose, let us define two notions, namely the \textit{variance function} and \textit{distance function}. In any given iteration $k$, define the \textit{variance function} $\sigma_k : \mathbb{R}^n \rightarrow \mathbb{R}_{\ge 0}$ as
\begin{equation} \label{eq:background:go:variance-function}
    \sigma_k(x) = \sqrt{ \sum_{i=1}^k{ v_{k,i}(x) \left( \hat{f}_k(x) - f_i \right)^2 } }.
\end{equation}
This function is able to induce exploratory behaviour by providing data-driven confidence intervals around the surrogate model. As a matter of fact, $\lim_{\delta \to 0} \sigma_k(x_i) = 0$, $i=1,\dots,k$ as no uncertainty affects points $x_i$ where $f$ has been evaluated exactly. Additionally, the \textit{distance function} $\zeta_k : \mathbb{R}^n \rightarrow \mathbb{R}_{\ge 0}$, defined by
\begin{equation} \label{eq:background:go:distance-function}
    \zeta_k(x) = \frac{2}{\pi} \arctan{ \left( \frac{1}{\sum_{i=1}^k{ w_i(x) }} \right) },
\end{equation}
approaches zero at sampled points (similarly to $\sigma_k$) but grows in between, pushing the GO algorithm towards unexplored regions. Given scalars $\lambda, \mu \ge 0$, the current dataset $\mathcal{D}_k$ and surrogate model $\hat{f}$, the overall acquisition function $\Lambda$ is then
\begin{equation} \label{eq:background:go:acquisition}
    \Lambda\left(\mathcal{D}_k, x\right) = \hat{f}_k(x) - \lambda \sigma_k(x) - \mu \Delta_{F_k} \zeta_k(x),
\end{equation}
where $\Delta_{F_k} = \max_{i=1,\dots,k}{f_i} - \min_{i=1,\dots,k}{f_i}$ is the range of observed samples $F_k$. As discussed in more detail in \cite{bemporad_global_2020}, the term $\hat{f}_k$ accounts for the exploitation of the samples $F_k$ collected so far, whereas the other two terms encourage exploration: $\sigma_k$ promotes high-uncertainty regions, and $\zeta_k$ unexplored ones.

However, similarly to most of the other classical acquisition functions in BO, the aforementioned acquisition function suffers from myopia, i.e., it is oblivious to the number of objective function evaluations left and does not take into account the evolution of the surrogate model due to future observations. Therefore, being essentially a one-step lookahead function, this acquisition function leads to myopic optimisation strategies. Fortunately, next is discussed dynamic programming, a tool that makes it possible to devise multi-step lookahead strategies that perform better and can balance the exploration-exploitation trade-off in an optimisation setting.

\subsection{Dynamic Programming and its Approximated Schemes}

Dynamic programming \citep{bertsekas_dynamic_2017} is a well-known method that provides us with a mathematical formulation to tackle optimal decision-making and control problems, in which we seek a policy (i.e., a sequence of control decisions) to control a (possibly stochastic) dynamical system in such a way as to minimise a given objective function. While not immediately straightforward, global optimal sequential estimation problems can also be cast as such, allowing us to leverage DP to address them formally in a nonmyopic way. For the sake of compactness, we limit ourselves to an introduction to DP and some of its approximated schemes.

\subsubsection{Dynamic Programming}

We consider a stochastic discrete-time system described by
\begin{equation}
    s_{k+1} = \mathcal{F}_k(s_k, a_k, w_k),
\end{equation}
where $s_k \in \mathcal{S}_k$ is the state at time step $k \in \left\{1,\dots,N\right\}$, $a_k \in \mathcal{A}_k$ the control action, and $w_k$ is a random process disturbance with probability distribution $\mathbb{P}_{w_k}$ with support $\mathcal{W}_k$. The function $\mathcal{F}_k : \mathcal{S}_k \times \mathcal{A}_k \times \mathcal{W}_k \rightarrow \mathcal{S}_{k+1}$ represents the (possibly nonlinear) dynamics that govern the underlying state transitions. A decision rule $\pi_k : \mathcal{S}_k \rightarrow \mathcal{A}_k$ is a mapping from states to actions, and a policy $\pi = \pi_1, \ldots, \pi_N$ is a sequence of rules. Moreover, let a stage-reward function $R_k : \mathcal{S}_k \times \mathcal{A}_k \rightarrow \mathbb{R}$ quantify the benefit of applying control action $a_k$ when the system is in state $s_k$.

Given these building blocks, the expected total return (or value function) over a finite horizon $N$ starting from time step $k=1$ with state $s_1$ and policy $\pi$ is given by\footnote{Note that a discount factor $\gamma$ is often included in \eqref{eq:background:dp:return} to discount future rewards in comparison to immediate ones. Here, however, it is omitted for simplicity as its effect on performance is not clear in the context of nonmyopic optimisation, as also mentioned in \cite{lam_bayesian_2016}.}
\begin{equation} \label{eq:background:dp:return}
    J_1^\pi(s_1) = \expval_{w_1,\dots,w_{N-1}}*{
        \sum_{k=1}^N{ R_k\bigl( s_k, \pi_k(s_k) \bigr) }
    },
\end{equation}
and the goal of DP is to find the optimal policy $\pi^\star$ in the space of admissible policies $\Pi$ that maximises this long-term return
\begin{maxi}
    { \scriptstyle \pi \in \Pi }{ J_1^\pi(s_1) }{}{ J_1^\star(s_1) = J_1^{\pi^\star}(s_1) = },
\end{maxi}
where, for simplicity, we assume the initial state $s_1$ to be exactly measurable and $\mathcal{S}_1 = \left\{s_1\right\}$. Following Bellman's principle of optimality, the problem can be solved using the DP recursive algorithm \citep{bertsekas_dynamic_2017}
\begin{equation} \label{eq:background:dp:recursive-algorithm}
    J^\star_{k}(s_k) = \max_{a_k \in \mathcal{A}_k}{
            R_k(s_k, a_k) +
            \expval_{w_k}[\Big]{J^\star_{k+1}\bigl(\mathcal{F}_k(s_k, a_k, w_k)\bigr)}
    },
    \quad \forall s_k \in \mathcal{S}_k,
\end{equation}
starting with $J^\star_{N+1}(s_{N+1}) = T(s_{N+1})$, and by moving backwards from $k=N$ to $k=1$. The optimal return is obtained at the last step as $J^\star_1(s_1)$, while the optimal policy can be found going forward, from $k=1$ to $k=N$, as one of the maximisers $a^\star_k = \pi^\star(s_k)$ of \eqref{eq:background:dp:recursive-algorithm}. The function $T : \mathcal{S}_{N+1} \rightarrow \mathbb{R}$ represents the terminal reward and, without loss of generality, is assumed to be zero in the rest of the paper.

Unfortunately, solving the DP problem exactly as described above is often only viable for problems with finite state and action spaces, as DP algorithms are known to suffer from the curse of dimensionality, especially as $N$ grows. The policy space $\Pi$ may also be infinite-dimensional, contributing to rendering the problem intractable. To address this issue, one needs to resort to approximate dynamic programming (ADP). Next, we briefly introduce two ADP techniques from the literature.

\subsubsection{Rollout}

Among the ADP methodologies, rollout stands out as a simple yet reliable sequential decision strategy to tackle the optimal control problem \citep{bertsekas_rollout_2020,bertsekas_multiagent_2021}. In short, starting from a base policy (e.g., a heuristic decided a priori by the user), rollout attempts to improve it via a one-step lookahead minimisation, thus emulating a single iteration of the policy iteration method. In mathematical terms, given the base policy $\mu = \mu_1,\dots,\mu_N$, for the current state $s_k$, the one-step lookahead rollout policy $\tilde{\mu}(s_k)$ can be computed as
\begin{argmaxi}
    { \scriptstyle a_k \in \mathcal{A}_k }{
        R_k(s_k, a_k) +
        \expval_{w_k}[\Big]{J_\mu\bigl(\mathcal{F}_k(s_k, a_k, w_k)\bigr)}
    }{ \label{eq:background:dp:rollout} }{ \tilde{\mu}(s_k) \in }.
\end{argmaxi}
Though similar to \eqref{eq:background:dp:recursive-algorithm} at first sight, \eqref{eq:background:dp:rollout} is meant to be solved online only for the current state $s_k$ and does not involve backwards computations over the entire state space. Instead, it under-approximates the optimal cost-to-go $J^\star_{k+1}$ with the heuristic base policy cost $J_\mu$, which is assumed much easier to compute since the base policy is known. Under some assumptions, it can be shown that the rollout policy enjoys an improvement property, i.e., $J_{\tilde{\mu}}(s_k) \ge J_\mu(s_k)$, $\forall s_k \in \mathcal{S}_k$ \citep[see][Section~6.4]{bertsekas_dynamic_2017}. The expectation operator in \eqref{eq:background:dp:rollout} is often the most expensive term to compute and, unless it can be formulated in an analytical expression, it is often addressed via approximations. There exist different ways of doing so \citep{bertsekas_dynamic_2017,bertsekas_rollout_2020}, and some of the most popular include Certainty Equivalence approaches (in which case, under additional assumptions, the expectation is removed from the objective), Monte Carlo sampling-based strategies (one of which is discussed next in \Cref{section:background:dp:ms}), and other cost-to-go approximations.

In this work, we employ a receding horizon control scheme, a variation of rollout formulation that uses a one-step minimisation base heuristic. Given a shorter horizon $H \le N$, the scheme only takes into account the next $h = \min{\left\{H, N-k+1\right\}}$ future sequential decisions, i.e., from time step $k$ to $k+h$, by solving the following stochastic optimal control problem:
\begin{maxi}
    { \scriptstyle \hat{a}_{k|k},\dots,\hat{a}_{k+h-1|k} }{
        R_t\left(\hat{s}_{k|k}, \hat{a}_{k|k}\right)
        + \expval_{w_k,\dots,w_{k+h-2}}*{
            \sum_{t=k+1}^{k+h-1}{
                R_t\left(\hat{s}_{t|k}, \hat{a}_{t|k}\right)
            }
        }
    }{ \label{eq:background:dp:rollout:mpc} }{}
    \addConstraint{ \hat{s}_{k|k} = s_k, }
    \addConstraint{
        \hat{s}_{t+1|k}
        = \mathcal{F}_t\left( \hat{s}_{t|k}, \hat{a}_{t|k}, w_t \right), \quad
    }{}{ t = k,\dots,k+h-1, }
    \addConstraint{ w_t \sim \mathbb{P}_{w_t}, }{}{ t = k,\dots,k+h-1, }
\end{maxi}
where $\hat{s}_{k|k},\dots,\hat{s}_{k+h|k}$ and $\hat{a}_{k|k},\dots,\hat{a}_{k+h-1|k}$ are the predictive state and action trajectories based on the information available at $k$. Then, only the first optimal control action $a_k = \hat{a}_{k|k}^\star$ is applied to the real system, leading to a new state $s_{k+1}$ at the next time step, at which point the procedure is repeated. In case the dynamics $\mathcal{F}_k$ are deterministic, computations for \eqref{eq:background:dp:rollout} and \eqref{eq:background:dp:rollout:mpc} simplify substantially as the evolution of the system is no longer aleatory (thus, there is no need to sample $w_t$), and the expectation can be elided from the objective. This formulation is heavily reminiscent of a single-shooting stochastic model predictive control (MPC) scheme, and interested readers are referred to \cite{rawlings_model_2017,mesbah_stochastic_2016} for more details. In the numerical experiments in this work, to make the rollout problem \eqref{eq:background:dp:rollout:mpc} tractable and solve it, the stochastic dynamics and expectation operators are approximated by replacing the disturbance signal $w_t$ with a sample drawn from $P_{w_t}$ at each predicted time step $t$. However, as further detailed in the next section, a more general and holistic sampling-based technique can be formulated to solve the control problem despite the presence of stochasticities.

\begin{remark}
    Note that perfect knowledge of the dynamics $\mathcal{F}_k$ is cardinal to guarantee the improvement property so as to predict exactly how the model evolves at the next time step. However, in practice, benefits in employing rollout can still be observed when only approximate models are available, as is the case in GO and BO \citep{yue_why_2020,bertsekas_rollout_2022}.
\end{remark}

\subsubsection{Multi-step Tree-based Optimisation}\label{section:background:dp:ms}

A paramount question is how to deal with the stochastic quantities in \eqref{eq:background:dp:rollout:mpc} in a non-deterministic case. A viable and tractable answer to it is to use sample-based techniques, which have been recently popularised in BO as well \citep{jiang_efficient_2020}.

Let $m_t \in \mathbb{N}_{>0}$, $t=k,\dots,k+h-2$, be the number of samples of the disturbance to be drawn at the prediction time step $t$. The core idea is then to replace the stochasticities in \eqref{eq:background:dp:rollout:mpc} with their sampled counterparts. Starting from the given initial state $s_k$, $m_k$ samples of the disturbance $\left\{w_k^{(1)}, \ldots, w_k^{(m_k)}\right\}$ are taken from $\mathbb{P}_{w_k}$ and, with the application of the first action $\hat{a}_{k|k}$, equally many next predicted states $\left\{\hat{s}_{k+1|k}^{(1)},\ldots,\hat{s}_{k+1|k}^{(m_k)}\right\}$ are obtained. For this first time step, the stage cost requires no approximation as it is deterministically found as $R_k\left(s_k,\hat{a}_{k|k}\right)$. In the next time step, for each state $\hat{s}_{k+1|k}^{(i_k)}$, $i_k = 1,\ldots,m_k$, a new set of $m_{k+1}$ disturbances $\left\{w_{k+1}^{(i_k,1)}, \ldots, w_{k+1}^{(i_k,m_{k+1})}\right\}$ is sampled according to $\mathbb{P}_{w_{k+1}}$, leading to approximating the next state evolution with the samples $\left\{\hat{s}_{k+2|k}^{(i_k, 1)},\ldots,\hat{s}_{k+2|k}^{(i_k,m_{k+1})}\right\}$ after applying $\hat{a}_{k+1|k}$. The stage cost for $k+1$ is then approximated as $\frac{1}{m_k} \sum_{i_k=1}^{m_k}{R_{k+1}\left(\hat{s}_{k+1|k}^{(i_k)}, \hat{a}_{k+1|k}\right)}$. The procedure is so repeated along the predicted horizon, creating a scenario tree-based approximation of the original problem \eqref{eq:background:dp:rollout:mpc}. The approach is visually summarised in \Cref{fig:background:dp:ms:tree}.
\begin{figure}
    \centering
    \tikzstyle{secondtolast} = [level distance=2cm, sibling distance=0.5cm]

\begin{tikzpicture}[
        grow'=right,
        level distance=4cm,
        sibling distance=1.75cm,
        sloped,
    ]
    \node (root) [circle, minimum width=3pt,fill, inner sep=0pt, label=west:{$s_k=\hat{s}_{k|k}$}] {}
    child {
        node (a) {$\hat{s}_{k+1|k}^{(1)}$} [secondtolast]
        child { node {$\dots$} }
        child { node {$\dots$} }
        child { node {$\dots$} }
        child [missing] {}
        child [missing] {}
        edge from parent node [above] {$w_k^{(1)}$}
    }
    child {
        node (b) {$\hat{s}_{k+1|k}^{(i_k)}$}
        child {
            node (aa) {$\hat{s}_{k+2|k}^{(i_k,1)}$} [secondtolast]
            child { node {$\dots$} }
            child { node {$\dots$} }
            child { node {$\dots$} }
            edge from parent node [above] {$w_{k+1}^{(i_k,1)}$}
        }
        child {
            node (bb) {$\hat{s}_{k+2|k}^{(i_k,i_{k+1})}$} [secondtolast]
            child { node {$\dots$} }
            child { node {$\dots$} }
            child { node {$\dots$} }
            edge from parent node [above] {$w_{k+1}^{(i_k,i_{k+1})}$}
        }
        child {
            node (cc) {$\hat{s}_{k+2|k}^{(i_k,m_{k+1})}$} [secondtolast]
            child { node (aaa) {$\dots$} }
            child { node {$\dots$} }
            child { node {$\dots$} }
            edge from parent node [above] {$w_{k+1}^{(i_k,m_{k+1})}$}
        }
        edge from parent node [above] {$w_k^{(i_k)}$}
    }
    child {
        node (c) {$\hat{s}_{k+1|k}^{(m_k)}$} [secondtolast]
        child [missing] {}
        child [missing] {}
        child { node {$\dots$} }
        child { node {$\dots$} }
        child { node {$\dots$} }
        edge from parent node [above] {$w_k^{(m_k)}$}
    };

    \path (a) -- (b) node [midway] {$\ldots$};
    \path (b) -- (c) node [midway] {$\ldots$};
    \path (aa) -- (bb) node [midway] {$\ldots$};
    \path (bb) -- (cc) node [midway] {$\ldots$};

    \def\Y{3.5}
    \draw [ultra thin] (root |- 0, \Y) -- (aaa |- 0, \Y);
    \foreach \nd in {root, a, aa, aaa} {
        \draw [ultra thin] (\nd |- 0, \Y) +(0mm,-1mm) -- +(0mm,1mm);
    }
    \path (root |- 0,\Y-0.3) -- (a |- 0,\Y-0.3) node [midway] {$\hat{a}_{k|k}$};
    \path (a |- 0,\Y-0.3) -- (aa |- 0,\Y-0.3) node [midway] {$\hat{a}_{k+1|k}$};
    \path (aa |- 0,\Y-0.3) -- (aaa |- 0,\Y-0.3) node [midway] {$\hat{a}_{k+2|k}$};
\end{tikzpicture}
    \caption{Depiction of the tree structure induced by the multi-step sampling of the disturbances, whereas the actions to be optimised remain constant within the same stage}
    \label{fig:background:dp:ms:tree}
\end{figure}
The resulting optimisation scheme is
\begin{maxi!}
    { \scriptstyle \hat{a}_{k|k},\dots,\hat{a}_{k+h-1|k} }{
        R_k\left(\hat{s}_{k|k}, \hat{a}_{k|k}\right)
        + \frac{1}{m_k} \sum_{i_k=1}^{m_k}{R_{k+1}\left(\hat{s}_{k+1|k}^{(i_k)}, \hat{a}_{k+1|k}\right)}
        + \dots
        \nonumber
    }{}{}
    \breakObjective{
        + \frac{1}{\prod_{\ell=k}^{k+h-2}{m_\ell}}
        \sum_{i_k=1}^{m_k}{
            \dots
            \sum_{i_{k+h-2}=1}^{m_{k+h-2}}{
                R_{k+h-1}\left(\hat{s}_{k+h-1|k}^{(i_k,\dots,i_{k+h-2})}, \hat{a}_{k+h-1|k}\right)
            }
        }
        \nonumber
    }
    \addConstraint{ \hat{s}_{k|k} = s_k, \tag{\theequation} \label{eq:background:dp:multi-step} }
    \addConstraint{
        \hat{s}_{t+1|k}^{(i_k,\dots,i_{t-1},i_t)}
        = \mathcal{F}_t\left(
            \hat{s}_{t|k}^{(i_k,\dots,i_{t-1})},
            \hat{a}_{t|k},
            w_t^{(i_k,\dots,i_{t-1},i_t)}
        \right),
        \nonumber
    }
    \addConstraint{ \hspace{3cm} i_t = 1,\dots,m_t, \ t = k,\dots,k+h-1. \nonumber }
\end{maxi!}
This formulation is advantageous because it circumvents the issue of stochastic dynamics and computations of expected values, and leaves us with a one-shot optimisation problem that does not involve nested maximisations and expectations, in contrast with \eqref{eq:background:dp:recursive-algorithm}. Under weak conditions, it can also be shown to lead to improvements with respect to a myopic approach \citep{jiang_efficient_2020}. On the other side, it is straightforward to notice that the approach lacks in scalability, especially when $m_t \gg 1$, since the number of scenarios in the tree grows exponentially with the horizon $h$, despite the number of primal variables growing only linearly.\footnote{In this regard, \cite{jiang_efficient_2020} shows how GPU-based automatic differentiation tools, such as \texttt{PyTorch} \citep{paszke_pytorch_2019}, can efficiently attenuate this drawback by leveraging massive parallelisation of the computations.}

Similarly to the rollout approach \eqref{eq:background:dp:rollout}, analogies to this sample-based approach can be found within the control community in the context of the MPC literature, e.g., \cite{blackmore_probabilistic_2010,schildbach_scenario_2014}. Moreover, notice that \eqref{eq:background:dp:multi-step} can be seen as a generalisation of \eqref{eq:background:dp:rollout}. In particular, when $m_t = 1$ and $w_t^{(1,\dots,1)} = \expval{w_t}$, \eqref{eq:background:dp:multi-step} regresses to a single-branch certainty equivalent solution to \eqref{eq:background:dp:rollout:mpc}.

\begin{remark}
    Again, it is important to stress that also in this formulation the performance of the resulting control policy is dependent on the quality of the prediction model $\mathcal{F}_k$, in case this is inexact. Likewise, the disturbance sampling strategy affects this performance, and different methods with different properties can be employed, e.g., Monte Carlo (MC), quasi-MC, or Gauss-Hermite (GH) sampling \citep{jiang_efficient_2020,lee_efficient_2020}.
\end{remark}

\section{Nonmyopic Global Optimisation} \label{section:nonmyopic-go}

In this section, we show how the GO sequential optimisation problem based on IDW and RBF regression can be formulated and tackled via the ADP techniques previously discussed, with the goal of crafting optimisation strategies that are systematically and structurally able to take future decisions into account and, thus, are nonmyopic. 

Given a surrogate model approximating the unknown objective function at unexplored query points, the additional necessary ingredients are
\begin{enumerate*}[label=\arabic*)]  
    \item a formulation of the dynamics describing how the surrogate model evolves when new information is made available and allowing to sample probable trajectories of such evolution, and
    \item a long-term objective goal to be minimised in the form of a reward function.
\end{enumerate*}
With these components in place, the GO sequential estimation problem can indeed be posed as a DP problem. By considering the iterations as the time axis, we can assign a state at a given iteration $k$ to the regression model to fully characterise it, and we can define the control action at time step $k$ as the next query point. By applying this control action, i.e., by observing the value of the unknown function at the query point, the state of the model is advanced to the next iteration $k+1$, i.e., the regression is updated to include the new information. With a cost associated with this state transition, DP and ADP algorithms can be deployed to optimise the selection of the next query points so as to minimise the cumulative costs incurred during the GO iterations. Therefore, this approach qualifies as nonmyopic as it takes into account the evolution of the underlying surrogate model and the cumulative costs incurred when choosing the sequence of points to be queried. Conversely, greedy GO strategies (e.g., see \Cref{section:background:globopt}) do not in general take into account the evolution of the underlying surrogate model for more than one iteration, and are considered myopic for this reason.

Whereas surrogate modelling has already been introduced and discussed in \Cref{section:background:globopt} in the form of IDW and RBF regression, the other two necessary ingredients, the surrogate models' dynamics and the reward function, are tackled in Sections \ref{section:nonmyopic-go:dynamics} and \ref{section:nonmyopic-go:reward}, respectively.

\subsection{Dynamics of Surrogate Models} \label{section:nonmyopic-go:dynamics}

The first ingredient to prepare in order to tackle GO problems via DP is a dynamical description of the two surrogate modelling approaches, namely IDW and RBF regression. The aim is to approximately predict and simulate, using only the information available at iteration $k$, how these models would evolve if new hypothetical observations were queried. This will then enable the use of DP techniques to improve upon the base GO sequential optimisation strategy.

\subsubsection{Via Inverse Distance Weighting}

Let us first tackle the case of IDW regression, which turns out to be the most straightforward. By leveraging the non-parametric nature of IDW, the model's state $s_k$ at each iteration $k$ is encoded solely by the corresponding dataset $\mathcal{D}_k$, which fully characterises the surrogate model $\hat{f}_k$, i.e., starting from any $\mathcal{D}_k$ we can use \eqref{eq:background:go:idw-regression} to fully construct $\hat{f}_k$. The control action $a_k$ is the next query point $x_{k+1}$ to be selected for evaluation. Thus, we have $\mathcal{S}_k = \left(\mathcal{X} \times \mathbb{R}\right)^k$ and $\mathcal{A}_k = \mathcal{X}$.

In formal terms, consider the initial state $\hat{s}_{k|k} = \mathcal{D}_k = \widehat{\mathcal{D}}_{k|k}$ and assume a sequence of control actions $a_t = x_{t+1}$, $t=k,\dots,k+h$, is given over a horizon of arbitrary length $h \ge 1$. It is possible to predict the evolution of the IDW regression as new data is added by approximating the value of the unknown objective function at the new query points (which cannot be directly observed as only information up to the current iteration $k$ can be used) as $f(x_{t+1}) \approx \hat{f}_{t|k}(x_{t+1})$. Therefore, as per \eqref{eq:background:go:recursive-dataset}, we define the next predicted state as $\hat{s}_{t+1|k} = \widehat{\mathcal{D}}_{t+1|k} = \hat{s}_{t|k} \cup \left\{\bigl( a_t, \hat{f}_{t|k}(a_t)\bigr)\right\}$, and the overall dynamics of the IDW model are represented as
\begin{equation} \label{eq:nmgo:approximate-dynamics:idw}
    \mathcal{F}_t(\hat{s}_{t|k}, a_t) = 
        \widehat{\mathcal{D}}_{t|k} \cup 
        \left\{\bigl( x_{t+1}, \hat{f}_{t|k}(x_{t+1})\bigr)\right\}, 
        \quad t = k,\dots,k+h.
\end{equation}
Obviously, this iterative procedure can only yield an approximate evolution of the regression model as additional hypothetical query-observation pairs are added during the sequential optimisation strategy. The reason is that it only exploits information available up to $k$, and in general $f(x_{t+1}) \neq \hat{f}_{t|k}(x_{t+1})$, especially in the earlier iterations when the dataset is small and less informative.

\subsubsection{Via Radial Basis Functions}

In case RBFs are used as surrogate model, an efficient representation of the regression evolution as new data is collected is less intuitive. Recall that the RBF regressor at iteration $k$ has coefficients $B_k$ that, together with points $X_k$, allow us to compute the estimate of objective $f$ at any $x \in \mathcal{X}$ according to \eqref{eq:background:go:rbf-regression}. Like IDW, the state of such a surrogate model can be defined solely as $s_k = \mathcal{D}_k$, but it is inconvenient to do so. In fact, this definition entails that each time new information is added as \eqref{eq:nmgo:approximate-dynamics:idw}, the linear system \eqref{eq:background:go:rbf-linear-system} has to be solved anew. It is clear then that as the iteration count $k$ increases each linear system becomes more and more computationally expensive to set up and solve.

To formulate an alternative, let us focus on how the regression coefficients change when a new query point is made available, i.e., the relationship between $B_k$ and $B_{k+1}$ when the new data point $\left( x_{k+1}, f_{k+1} \right)$ is added to the dataset. Assume $B_k = M_k^{-1} F_k$ has been previously computed, e.g., as per \Cref{section:background:globopt}. It is then possible to write $M_{k+1}^{-1}$ in an incremental and recursive way as the blockwise inverse \citep[see][Section~28.2]{cormen_introduction_2009}
\begin{equation} \label{eq:nmgo:approximate-dynamics:rbf:blockwise-inverse}
    M_{k+1}^{-1} =
    \begin{bmatrix}
        M_k                  & \Phi_k(x_{k+1})                                              \\
        \Phi_k(x_{k+1})^\top & \phi\left(\varepsilon d\left(x_{k+1}, x_{k+1}\right) \right)
    \end{bmatrix}^{-1} \\
    = \frac{1}{c}
    \begin{bmatrix}
        c M_k^{-1} + L L^\top & - L \\
        - L^\top              & 1
    \end{bmatrix},
\end{equation}
where
\begin{align}
    L ={}& M_k^{-1} \Phi_k(x_{k+1}),                                                                                     \\
    c ={}& \phi\left(\varepsilon d\left(x_{k+1}, x_{k+1}\right) \right) - \Phi_k(x_{k+1})^\top M_k^{-1} \Phi_k(x_{k+1}).
\end{align}
In this context, $c \in \mathbb{R}$ is also known as the Schur complement. It is assumed here that $c \neq 0$.\footnote{For simplicity, in those cases where this assumption fails, we fall back to solving the full linear system \eqref{eq:background:go:rbf-linear-system} instead. However, more refined incremental strategies that do not need such an assumption could be employed, e.g., incremental singular value decomposition \citep{brand_fast_2006}. Alternatively, the sequential decision algorithm may be appropriately penalised for or constrained from selecting any $x_{k+1}$ for which $\lvert c \rvert < \xi $, for some small $\xi > 0$.} Hence, the blockwise inverse \eqref{eq:nmgo:approximate-dynamics:rbf:blockwise-inverse} allows us to efficiently get the new coefficients as $B_{k+1} = M_{k+1}^{-1} \begin{bmatrix} F_k \\ f_{k+1} \end{bmatrix}$, without having to solve the full system of equations \eqref{eq:background:go:rbf-linear-system}. For convenience of notation, let us define the function $\mathcal{M}_k : \mathbb{R}^{k \times k} \times \mathbb{R}^{n} \rightarrow \mathbb{R}^{k+1 \times k+1}$ as a compact shorthand for this procedure, i.e., 
\begin{equation} \label{eq:nmgo:approximate-dynamics:rbf}
    M_{k+1}^{-1} = \mathcal{M}_k \left(M_k^{-1}, x_{k+1}\right).
\end{equation}
Therefore, in order to exploit the recursive blockwise inversion scheme, in comparison to the case of IDWs, the state of the RBF regression at iteration $k$ must be enhanced with the current inverse of the interpolation matrix, i.e., $\mathcal{S}_k = \left(\mathcal{X} \times \mathbb{R}\right)^k \times \mathbb{R}^{k\times k}$, while the action space remains unchanged, i.e., $\mathcal{A}_k = \mathcal{X}$. In summary, given the initial state $\hat{s}_{k|k} = \left( \widehat{\mathcal{D}}_{k|k}, \widehat{M}_{k|k}^{-1} \right)$ and the sequence of control actions $a_t = x_{t+1}$, $t=k,\dots,k+h$, the dynamics of the RBF regression model are given by
\begin{equation}
    \mathcal{F}_t\left(\hat{s}_{t|k}, a_t\right) = \biggl(
        \widehat{\mathcal{D}}_{t|k} 
        \cup \left\{\bigl( x_{t+1}, \hat{f}_{t|k}(x_{t+1}) \bigr)\right\}, \
        \mathcal{M}_t \left(\widehat{M}_{t|k}^{-1}, x_{t+1}\right) 
    \biggr),
    \quad t = k,\dots,k+h.
\end{equation}
As before, we remark these dynamics are an approximation due to the fact that, as we move along the prediction horizon from $k$ to $k+h$, we are using the surrogate model to simulate the value of the objective function at the new query points instead of measuring its real value, and in general we have $f(x_{t+1}) \neq \hat{f}_{t|k}(x_{t+1})$.

\subsubsection{Sampling}

So far, the dynamics for IDW and RBF regression have been discussed. These surrogate models evolve deterministically because their estimates $\hat{f}_k$ are intrinsically so. However, in order to apply the multi-step tree-based approach \eqref{eq:background:dp:multi-step}, we need to be able to sample possible trajectories of the regressor dynamics when the objective function is approximated at new query points. Note that, contrarily to IDW- and RBF-based GO, in BO such a property is readily available thanks to the employment of GP models. In fact, a GP regressor offers a probabilistic prediction of the unknown objective function at an unexplored point in the form of a Gaussian posterior distribution. This distribution can then be sampled, thus allowing the simulation of different possible evolutions of the regressor.

In this work, we opt to enhance our IDW and RBF predictors with similar behaviour as GPs by inducing Gaussian posterior distributions via the variance function $\sigma_k(x)$ in \eqref{eq:background:go:variance-function}. This function was introduced in \Cref{section:background:globopt}, in the context of the myopic acquisition function, as a heuristic able to provide confidence bounds to induce exploration towards regions of the space $\mathcal{X}$ where the regression prediction is less trustworthy. However, the same function can also be exploited to sample likely estimates of the unknown objective based on IDW/RBF surrogate models, thus rendering these more amenable to the formulation in \eqref{eq:background:dp:multi-step}.

Consider an IDW or RBF regression model fitted on $\mathcal{D}_k$, at iteration $k$. The stochastic estimate of the unknown objective $f$ at $x_{k+1}$ is now the random variable $\hat{f}^\text{s}_k$ such that
\begin{equation} \label{eq:nmgo:dynamics:sampling:posterior}
    \hat{f}^\text{s}_k \left(x_{k+1}\right) \sim \mathcal{N}\left( \hat{f}_k \left(x_{k+1}\right), \sigma_k^2(x_{k+1}) \right),
\end{equation}
where $\hat{f}_k \left(x_{k+1}\right)$ follows from either \eqref{eq:background:go:idw-regression} or \eqref{eq:background:go:rbf-regression}, and $\sigma_k$ has been introduced in \eqref{eq:background:go:variance-function}. Thanks to this definition of $\hat{f}^\text{s}_k$, we can then sample its distribution to get different sample-based state trajectories of the regressor. We stress here that this estimated distribution is heuristic, contrarily to GPs where typically suitable assumptions are put in place to guarantee that such posterior distributions are theoretically sound. 

\subsection{Reward Function}  \label{section:nonmyopic-go:reward}

The second ingredient before exploiting the ADP formulation \eqref{eq:background:dp:multi-step} to craft a nonmyopic GO strategy is to define a suitable reward function. In this work, we choose to define such reward (both for the IDW and RBF setting) as
\begin{equation} \label{eq:nmgo:reward}
    \begin{aligned}
        - R_k(s_k, a_k) ={}& \Lambda^\text{s}\left(\mathcal{D}_k, x_{k+1}\right) \\
        \coloneqq{}& \expval_{\hat{f}^\text{s}_k}[\big]{ \Lambda\left(\mathcal{D}_k, x_{k+1}\right) } \\
        ={}& \hat{f}_k(x_{k+1}) - \lambda \expval_{\hat{f}^\text{s}_k}[\big]{\sigma_k(x_{k+1})} - \mu \Delta_{F_k} \zeta_k(x_{k+1}),
    \end{aligned}
\end{equation}
where, instead of using the deterministic prediction $\hat{f}_k$, we advocate to the use of the stochastic posterior $\hat{f}^\text{s}_k$ and the computation of the corresponding reward expectation. Since $\zeta_k$ does not depend on the prediction and $\expval_{\hat{f}^\text{s}_k}*{\hat{f}^\text{s}_k(x_{k+1})} = \hat{f}_k(x_{k+1})$ by definition, we are left with just the computation of the expected value of the variance function. No simple closed-form solution to this exists but, since the expectation is over a normally distributed variable, one can approximate it up to an arbitrary precision via GH quadrature \citep{abramowitz_handbook_1972}, i.e.,
\begin{equation}
    \expval_{\hat{f}^\text{s}_k}[\big]{\sigma_k(x_{k+1})} \approx \sum_{j=1}^{q}{
        \xi_j \sqrt{ \sum_{i=1}^k{ v_{k,i}(x) \left( z_j - f_i \right)^2 } }
    },
\end{equation}
where $q \in \mathbb{N}_{>0}$ is the number of sample points, $z_j \in \mathbb{R}$, $j=1,\dots,q$, are appropriately spaced (depending on $q$) samples of $\hat{f}^\text{s}_k\left(x_{k+1}\right)$ and $\xi_j \in \mathbb{R}$, $j=1,\dots,q$, are the associated weights found as
\begin{equation}
    \xi_j = \frac{ 2^{q-1} q! \sqrt{\pi} }{q^2 H^2_{q-1}(\varrho_j)^2 },
\end{equation}
with $\varrho_j$, $j=1,\dots,q$, the roots of the Hermite polynomial $H_q : \mathbb{R} \rightarrow \mathbb{R}$. Lastly, note that a negative sign has been included in \eqref{eq:nmgo:reward} to flip the optimisation direction because, whereas $R_k$ is maximised by convention, $\Lambda$ must be minimised.

\subsection{Nonmyopic Acquisition Problems and Global Optimisation}

In summary, we have shown how the surrogate model (be it IDW or RBF) can be regarded as a dynamical system and how it is possible to approximately predict its evolution during the optimisation strategy when an action is taken, i.e., a new query point is selected. Furthermore, we have enabled sampling this evolution of such models by defining a heuristic distribution of the objective estimate (so as to be able to generate different possible state trajectories), as well as modified the vanilla GO myopic acquisition function to take into account this new probabilistic approach.

Inspired by the multi-step tree-based approach \eqref{eq:background:dp:multi-step}, we propose the following nonmyopic acquisition methods that combine and take advantage of the aforementioned modifications. Compared to the original myopic acquisition \eqref{eq:background:go:acquistion-problem}, the proposed ones are predictive, and thus they are able to carry out nonmyopic decisions that foresee and take into consideration the evolution of the surrogate model.

For the case of IDW regression, we define at iteration $k$ the following nonmyopic acquisition problem:
\begin{mini!} 
    { \scriptstyle \hat{x}_{k+1|k},\dots,\hat{x}_{k+h|k} }{
        \Lambda^\text{s}\left(\widehat{\mathcal{D}}_{k|k}, \hat{x}_{k+1|k}\right)
        + \frac{1}{m_k} \sum_{i_k=1}^{m_k}{\Lambda^\text{s}\left(\widehat{\mathcal{D}}_{k+1|k}^{(i_k)}, \hat{x}_{k+2|k}\right)}
        + \dots
        \nonumber \label{eq:nmgo:acqfun:idw:objective}
    }{ \label{eq:nmgo:acqfun:idw} }{}
    \breakObjective{
        + \frac{1}{\prod_{\ell=k}^{k+h-2}{m_\ell}}
        \sum_{i_k=1}^{m_k}{
            \dots
            \sum_{i_{h-1}=1}^{m_{k+h-2}}{
                \Lambda^\text{s}\left(\widehat{\mathcal{D}}_{k+h-1|k}^{(i_k,\dots,i_{k+h-2})}, \hat{x}_{k+h|k}\right)
            }
        }
    }
    \addConstraint{ \widehat{\mathcal{D}}_{k|k} = \mathcal{D}_k, \label{eq:nmgo:acqfun:idw:constraint:initial-dataset}}
    \addConstraint{
        \widehat{\mathcal{D}}_{t+1|k}^{(i_k,\dots,i_{t-1},i_t)}
        = \widehat{\mathcal{D}}_{t|k}^{(i_k,\dots,i_{t-1})}
        \cup \biggl\{\left(
            \hat{x}_{t+1|k}, \hat{f}_{t|k}^{(i_k,\dots,i_{t-1},i_t)} 
        \right)\biggr\},
    }
    \addConstraint{ \hspace{3cm} i_t = 1,\dots,m_t, \ t = k,\dots,k+h-1, \nonumber }
    \addConstraint{ 
        \hat{x}_{t+1|k} \in \mathcal{X}, 
        \hspace{3.567cm} t = k,\dots,k+h-1,
        \label{eq:nmgo:acqfun:idw:constraint:constrained-actions}
    }
\end{mini!}
and take as the next query point its first optimal action, i.e., $x_{k+1} = \hat{x}_{k+1|k}^\star$. Since \eqref{eq:nmgo:acqfun:idw} is nonlinear, we break ties equally likely in case of multiple equivalent local minima. Note that the samples of the dynamics evolution $\hat{f}_t^{(i_k,\dots,i_t)}$, $i_t=1,\dots,m_t$, are drawn from $\hat{f}^\text{s}_t \left(\hat{x}_{t+1|k}\right) \sim \mathcal{N}\left( \hat{f}_t \left(\hat{x}_{t+1|k}\right), \sigma_t^2 \left(\hat{x}_{t+1|k}\right) \right)$. To preserve differentiability through this sampling step for the sake of computing the sensitivity of the objective \eqref{eq:nmgo:acqfun:idw:objective} w.r.t. the primal variables, the well-known re-parametrisation trick is used \citep{kingma_autoencoding_2013}, i.e., samples are drawn as $\hat{f}^\text{s}_t \left(\hat{x}_{t+1|k}\right) = \hat{f}_t \left(\hat{x}_{t+1|k}\right) + \sigma_t \left(\hat{x}_{t+1|k}\right) z$, with $z \sim \mathcal{N}(0,1)$.

Slightly more convoluted is the case for RBF regression, where the additional state entry leads to another set of constraints to be included in the optimisation, leading us to the following problem:
\begin{mini}
    { \scriptstyle \hat{x}_{k+1|k},\dots,\hat{x}_{k+h|k} }{ \eqref{eq:nmgo:acqfun:idw:objective} }{ \label{eq:nmgo:acqfun:rbf} }{}
    \addConstraint{
        \eqref{eq:nmgo:acqfun:idw:constraint:initial-dataset}-\eqref{eq:nmgo:acqfun:idw:constraint:constrained-actions}, 
    }
    \addConstraint{ \widehat{M}_{k|k}^{-1} = M_k^{-1}, }
    \addConstraint{
        \widehat{M}_{t+1|k}^{-1}
        = \mathcal{M}_t \left( \widehat{M}_{t|k}^{-1}, \hat{x}_{t+1|k} \right), \quad}{}{t = k,\dots,k+h-1.}
\end{mini}
One can see that, as they do not depend on the samples $\hat{f}_t^{(i_k,\dots,i_t)}$, the predicted interpolation matrices $\widehat{M}_{t|k}^{-1}$, $t=k,\dots,k+h$ need not be sampled, as opposed to $\widehat{\mathcal{D}}_{t|k}^{(i_k,\dots,i_t)}$. All other considerations from \eqref{eq:nmgo:acqfun:idw} apply also to \eqref{eq:nmgo:acqfun:rbf}.

\Cref{algo:nmgo} summarises the proposed nonmyopic methodology. Compared to its myopic counterpart, two major improvements are introduced. In particular, the vanilla myopic acquisition problem \eqref{eq:background:go:acquistion-problem} is replaced with the nonmyopic one in order to combat the greediness of the standard GO formulation and to improve upon its performance. Note that for $h=1$, i.e., the last iteration, we fall back to solving the (stochastic) myopic problem as there is no need to have a predictive strategy because the algorithm is about to terminate; equivalently, one could notice that for $h=1$ problems \eqref{eq:nmgo:acqfun:idw} and \eqref{eq:nmgo:acqfun:rbf} degenerate to the minimisation in \cref{algo:nmgo:last-iter}. On top of this first modification, instead of training in each iteration $k$ the IDW or RBF regression on the whole dataset $\mathcal{D}_k$ from scratch, we also propose to take again advantage of the dynamical formulation of these surrogate models to partially fit them only on the new data $\left( x_{k+1}, f_{k+1} \right)$. This is, in general, more computationally efficient, especially in the case of RBF regression where determining solutions to the linear system \eqref{eq:background:go:rbf-linear-system} can become computationally expensive for large $k$.
\begin{algorithm}
    \caption{Nonmyopic GO Algorithm} \label{algo:nmgo}
    \begin{algorithmic}[1]
    \Require initial dataset $\mathcal{D}_{N_0}$; search space $\mathcal{X}$; maximum number of iterations $N$; horizon $H$; number of samples $m_k$, $k=N_0,\dots,N$ ($m_k=1$, if rollout).
    \For{$k = N_0 \to N$}
        \If{$k = N_0$}
            \State Fit initial regression model $\hat{f}_k$ to $\mathcal{D}_{N_0}$ via \eqref{eq:background:go:idw-regression} or \eqref{eq:background:go:rbf-regression}
        \Else
            \State Partially fit regression model $\hat{f}_k$ to new data via \eqref{eq:background:go:recursive-dataset} and, if using RBF, \eqref{eq:nmgo:approximate-dynamics:rbf}
        \EndIf
        \State $h \gets \min{\left\{H, N-k+1\right\}}$ \Comment{\textit{compute current horizon}}
        \If{$h > 1$}
            \State Get new query point $x_{k+1} = \hat{x}_{k+1|k}^\star$ by solving \eqref{eq:nmgo:acqfun:idw} or \eqref{eq:nmgo:acqfun:rbf}
        \Else
            \State Get new query point by $\min_{ x\in\mathcal{X} }{ \Lambda^\text{s}\left(\mathcal{D}_N, x\right) }$ \Comment{\textit{last iteration}} \label{algo:nmgo:last-iter}
        \EndIf
        \State $f_{k+1} \gets f(x_{k+1})$  \Comment{\textit{observe value at query point}}
    \EndFor
    \State \Return $\max_{i=1,\dots,N+1} x_i$
    \end{algorithmic}
\end{algorithm}

\section{Numerical Experiments} \label{section:numerical-experiments}

In this section, we implement the proposed nonmyopic GO optimisation strategies \eqref{eq:nmgo:acqfun:idw}-\eqref{eq:nmgo:acqfun:rbf} and assess the resulting performance on two different experiments, namely, on a collection of several benchmark synthetic and real-world functions from the optimisation and machine learning literature, and on the tuning of a controller for a chemical control task. Our implementation builds on top of \texttt{BoTorch} \citep{balandat_botorch_2020}, a Python framework for GO/BO that leverages \texttt{PyTorch} \citep{paszke_pytorch_2019} for massive GPU-based parallelisation and differentiation of its computations. Source code and simulation results are made available in the following repository: \url{https://github.com/FilippoAiraldi/global-optimization}.

In the benchmarking and MPC numerical experiments, the search space $\mathcal{X}$ is a hyperrectangle, i.e., $\mathcal{X} = \left\{ x \in \mathbb{R}^n \;\middle|\; \underline{x}_i \le x_i \le \overline{x}_i, \ i=1,\dots,n \right\}$ for some problem-specific lower and upper bounds $\underline{x}, \overline{x} \in \mathbb{R}^n$. Due to their high nonlinearity, the nonmyopic optimisation problems are solved via L-BFGS-B \citep{byrd_limited_1995,zhu_algorithm_1997}, a quasi-Newton method designed to tackle problems with box-constrained variables, and with multistart to avoid convergence to very suboptimal local minima. As suggested in \cite{bemporad_global_2020}, the coefficients of the acquisition function are scaled with the size of the search space as $\lambda = n^{-1}$ and $\mu = 0.5 n^{-1}$. In \eqref{eq:background:go:idw-weights} $\delta = 10^{-12}$, and a threshold of $10^{-8}$ on the smallest singular values is imposed (see \Cref{section:background:globopt} and \cite{bemporad_global_2020}, Section~3.2, for more details on this). For those problems that are regressed via RBF, we estimate $\varepsilon$ every 10 iterations via a 5-fold cross validation on the current dataset, where candidates are generated randomly from a log-normal distribution. Lastly, the number of samples to approximate the expectation in \eqref{eq:nmgo:reward} via GH quadrature is fixed as $q=16$. This value was tuned to strike a balanced trade-off between approximation accuracy and computational performance.

Each numerical experiment is run on a server with Python 3.11.3 and equipped with 16 AMD EPYC 7252 (3.1 GHz) processors, 252GB RAM, and 4 NVIDIA RTX 3090 GPUs.

\subsection{Synthetic and Real-world Problems} \label{section:numerical-experiments:benchmarks}

We test the proposed nonmyopic approaches on a collection of standard synthetic benchmark functions \citep{surjanovic_virtual_2024,jamil_literature_2013,brochu_active_2007} as well as on a set of real-world hyperparameter tuning problems \citep{snoek_practical_2012,wang_maxvalue_2017,malkomes_automating_2018,jiang_efficient_2020}. \Cref{tab:numerical:synthetic-and-real:list} reports all the functions, including their search space dimension and bounds. Each is solved in two stages: first, $N_0 = 2n$ initial random points are drawn uniformly at random to warm the surrogate models up; then, the main sequential algorithm is run to pursue the function's optimum. To combat the influence of randomness, each experiment is repeated 30 times with different seeds, while the rest of the other hyperparameters is kept the same.
\begin{table}[t]
    \centering
    \caption{Synthetic benchmarks and real-world hyperparameter tuning problems}
    \begin{tabular}{@{}llcccc@{}}
        \toprule
        Group & Name                              & $n$       & $\underline{x}$                      & $\overline{x}$                                    \\ 
        \midrule \multirow{13}*{Synthetic}
              & Ackley                            & 2         & -32.767                              & 32.767                                            \\
              & Adjiman                           & 2         & -1                                   & $\left[ 2, 1 \right]$                             \\
              & Beale                             & 2         & -4.5                                 & 4.5                                               \\
              & Bohachevsky                       & 2         & -100                                 & 100                                               \\
              & Brochu (2d, 4d, 6d)               & 2,\,4,\,6 & 0                                    & 1                                                 \\
              & Bukin                             & 2         & $\left[ -15, -3 \right]$             & $\left[ -5, 3 \right]$                            \\
              & Camel hump (3d)                   & 2         & -5                                   & 5                                                 \\
              & Camel hump (6d)                   & 2         & $\left[ -3, -2 \right]$              & $\left[ 3, 2 \right]$                             \\
              & Dixon-Price (2d, 4d)              & 2,\,4     & -10                                  & 10                                                \\
              & Drop-wave                         & 2         & -5.12                                & 5.12                                              \\
              & Egg holder                        & 2         & -512                                 & 512                                               \\
              & Griewank                          & 2         & -600                                 & 600                                               \\
              & Hartman (3d, 6d)                  & 3,\,6     & 0                                    & 1                                                 \\
              & Himmelblau                        & 2         & -5                                   & 5                                                 \\
              & Levy (2d, 4d, 6d)                 & 2,\,4,\,6 & -10                                  & 10                                                \\
              & Powell                            & 4         & -4                                   & 5                                                 \\
              & Rastrigin                         & 2         & -5.12                                & 5.12                                              \\
              & Rosenbrock                        & 8         & -5                                   & 10                                                \\
              & Shubert                           & 2         & -5.12                                & 5.12                                              \\
              & Step 2                            & 5         & -100                                 & 100                                               \\
              & Styblinski-Tang                   & 5         & -5                                   & 5                                                 \\ 
        \cmidrule{2-5} \multirow{7}*{Real-world}
              & Cosmological                      & 9         & 0                                    & 1                                                 \\
              & LDA                               & 3         & $\left[0.5, 0, 0\right]$             & $\left[1, 10, 14\right]$                          \\
              & LogReg                            & 4         & $\left[ 0, 0, 4.32, 2.32\right]$     & $\left[10, 1, 11.32, 11.32\right]$                \\
              & NN Boston                         & 4         & 0                                    & 1                                                 \\
              & NN Cancer                         & 4         & 0                                    & 1                                                 \\
              & Robot push (3d, 4d)               & 3,\,4     & 0                                    & 1                                                 \\
              & SVM                               & 3         & $\left[ -3.32, -3.32, -13.29\right]$ & $\left[ 19.93, 2.32, -3.32\right]$                \\ \botrule
    \end{tabular}
    \label{tab:numerical:synthetic-and-real:list}
    \footnotetext{Note: a scalar bound is reported if it applies to all dimensions}
\end{table}

The rollout and the multi-step approach are both tested with different horizon lengths $H$. The former is tested with a lookahead of up to five steps, while the latter is tested only up to four due to computational complexity. For the multi-step approach, the numbers of samples at iteration $k$ are set to $m_k=10$, $m_{k+1}=5$, $m_{k+2}=3$, and are kept constant across iterations. For rollout, obviously we have that $m_t=1$, $t=k,\dots,k+h-2$, with $h \in \{2,\dots,5\}$. The sampling of uncertain observations is carried out according to the distribution \eqref{eq:nmgo:dynamics:sampling:posterior} for both methodologies. This is achieved, in line with the state of the art \citep{jiang_efficient_2020}, randomly via MC sampling of the normal distribution (in particular, via the Quasi-MC Box-Muller method \citep{pages_numerical_2018}).

To quantitatively appreciate the performance of the algorithms, the optimality gap $G$, defined as
\begin{equation}
    G = \frac{\displaystyle\min_{i=1,\dots,N_0}{ f_i } - \min_{i=1,\dots,N+1}{ f_i } }{\displaystyle\min_{i=1,\dots,,N_0}{ f_i } - f(x^\star) },
\end{equation}
is employed as a metric that measures the improvement in the unknown objective function from the initial iteration $N_0$ to the final iteration $N$, normalised by the maximum achievable reduction. Note that, the closer the gap is to $1$, the less suboptimal the convergence point is\footnote{This holds true except in the extremely unlikely case in which the global optimal point is already found in the first $N_0$ random iterations, i.e., $\exists i \in \{1,\dots,N_0\} \, : \, f_i = f(x^\star)$. If so, this metric is no longer well-defined.}. 

The results on the mean and median gap, obtained with the aforementioned settings, are reported in \Cref{tab:numerical:synthetic-and-real:gap}. The functions' groups and names follow from \Cref{tab:numerical:synthetic-and-real:list}, while methods' names are abbreviated to include strategy and horizon, e.g., \textit{R-3} and \textit{MS-3} correspond to a rollout and multi-step tree-based approach with horizon $H=3$, respectively. For the sake of comparison, we also include results for 
\begin{itemize}
    \item the myopic strategy \citep{bemporad_global_2020} upon which we have built our proposed approach, forming our reference baseline (\textit{B})
    \item a classical myopic BO method, with EI as acquisition function (\textit{EI})
    \item the nonmyopic rollout BO method by \cite{jiang_efficient_2020}, which has been shown to outcompete other approaches such as GLASSES \citep{gonzalez_glasses_2016} and BINOCULARS \citep{jiang_binoculars_2020}, with EI as stage-wise reward (\textit{EIR-H}, with \textit{H} the horizon length).
\end{itemize}
The best mean/median result in each problem is highlighted in {\color{blue}\textbf{blue bold}}, whereas results that are not significantly worse than the best (under the one-sided Wilcoxon signed-rank test with $p=5\%$) are reported in {\color{purple}\textit{purple italics}}. Alongside this table, we also provide \Cref{fig:numerical:itertime-vs-gap}, which summarises the average performance (in terms of optimality gap) versus the mean computation time per iteration for each simulated method.
\begin{figure}
    \centering
    \begin{tikzpicture}
    \pgfplotstableread{data/benchmarking/itertime-vs-gap.dat}\data
    \pgfmathsetmacro{\myymin}{0.645}
    \pgfmathsetmacro{\myymax}{0.86}

    \begin{axis}[
        name=plot-background,
        ticks=none,
        grid=none,
        axis background/.style={draw=none},
        axis line style={draw=none},
        width=\axisdefaultwidth,
        height=\axisdefaultheight,
        xlabel=Seconds per iteration (s),
        xlabel shift=0.68cm,
        ylabel={Optimality gap},
        ylabel shift=0.85cm,
    ]
    \end{axis}

    \begin{semilogxaxis}[
        name=plot-left,
        at={($(plot-background.west)$)},
        anchor=west,
        width=0.3\textwidth,
        height=\axisdefaultheight,
        xmin=0.55, xmax=2.55,
        xtick={1},
        minor xtick={0.4,0.5,0.6,0.7,0.8,0.9,2,3},
        axis y line*=left,
        ymin=\myymin, ymax=\myymax,
        visualization depends on={\thisrow{anchor} \as \thisanchor},
        nodes near coords style={font=\small\itshape, anchor=\thisanchor},
    ]
        \pgfplotsinvokeforeach{1,4}{
            \addplot [
                C#1,
                only marks,
                mark=*,
                mark options={draw=white},
                restrict expr to domain={\thisrow{type}}{#1:#1},
                point meta=explicit symbolic,
                nodes near coords,
                error bars,
            ] table [x=time, y=gap, x error=time-err, y error=gap-err, meta=method] {\data};
        }
    \end{semilogxaxis}

    \begin{semilogxaxis}[
        name=plot-right,
        at={($(plot-background.east)$)},
        anchor=east,
        width=0.86\textwidth,
        height=\axisdefaultheight,
        xmin=4, xmax=55,
        xtick={10},
        minor xtick={3,4,5,6,7,8,9,20,30,40,50,60},
        ymin=\myymin, ymax=\myymax,
        axis y line*=right,
        yticklabels={},
        ytick style={draw=none},
        visualization depends on={\thisrow{anchor} \as \thisanchor},
        nodes near coords style={font=\small\itshape, anchor=\thisanchor},
        after end axis/.code={
            \draw (rel axis cs:0.02,0) +(-1mm,-2mm) -- +(1mm,2mm);
            \draw (rel axis cs:0.03,0) +(-1mm,-2mm) -- +(1mm,2mm);
        },
    ]
        \pgfplotsinvokeforeach{0,2,3}{
            \addplot [
                C#1,
                only marks,
                mark=*,
                mark options={draw=white},
                restrict expr to domain={\thisrow{type}}{#1:#1},
                point meta=explicit symbolic,
                nodes near coords,
                error bars,
            ] table [x=time, y=gap, x error=time-err, y error=gap-err, meta=method] {\data};
        }
    \end{semilogxaxis}

\end{tikzpicture}
    \caption{Average gap (aggregated over all problems) versus the mean time per iteration $\pm$ the corresponding standard error of the mean for various methods with different horizon lengths. Note that the time axis is in logarithmic scale and has a small discontinuity}
    \label{fig:numerical:itertime-vs-gap}
\end{figure}
\afterpage{\begin{landscape}
	\setlength{\tabcolsep}{3.1pt}
	\setlength{\LTcapwidth}{7.5in}
	\begin{longtable}[c]{@{}ll|c|cccc|ccc|c|cccc@{}}
		\caption{Mean and median final optimality gap on the synthetic and real functions for all methods, with best results in {\color{blue}\textbf{blue bold}} and results not significantly worse than the best in {\color{purple}\textit{purple italics}}}
        \label{tab:numerical:synthetic-and-real:gap} \\
		\toprule 
		Name                                                  &        & \textit{B}                     & \textit{R-2}                   & \textit{R-3}                   & \textit{R-4}                   & \textit{R-5}                   & \textit{MS-2}                  & \textit{MS-3}                  & \textit{MS-4}                  & \textit{EI}                    & \textit{EIR-2}                 & \textit{EIR-3}                 & \textit{EIR-4}                 & \textit{EIR-5}                 \\
		\midrule
		\endfirsthead
		\toprule 
		Name                                                  &        & \textit{B}                     & \textit{R-2}                   & \textit{R-3}                   & \textit{R-4}                   & \textit{R-5}                   & \textit{MS-2}                  & \textit{MS-3}                  & \textit{MS-4}                  & \textit{EI}                    & \textit{EIR-2}                 & \textit{EIR-3}                 & \textit{EIR-4}                 & \textit{EIR-5}                 \\
		\midrule
		\endhead
		\botrule 
		\multicolumn{15}{l}{\textsuperscript{\dag}An RBF surrogate model was used instead of IDW}
		\endlastfoot
		\multirow{2}*{Ackley}                                 & mean   & 0.985                          & 0.979                          & 0.975                          & 0.963                          & 0.954                          & {\color{purple}\textit{0.995}} & {\color{blue}\textbf{0.998}}   & {\color{purple}\textit{0.997}} & 0.534                          & 0.883                          & 0.862                          & 0.833                          & 0.787                          \\
															  & median & 0.998                          & 0.997                          & 0.994                          & 0.993                          & 0.983                          & {\color{blue}\textbf{0.998}}   & {\color{purple}\textit{0.998}} & {\color{purple}\textit{0.998}} & 0.870                          & 0.973                          & 0.943                          & 0.945                          & 0.841                          \\
		\multirow{2}*{Adjiman}                                & mean   & 0.627                          & {\color{purple}\textit{0.935}} & {\color{purple}\textit{0.945}} & {\color{purple}\textit{0.914}} & {\color{purple}\textit{0.885}} & {\color{purple}\textit{0.938}} & {\color{purple}\textit{0.946}} & {\color{purple}\textit{0.967}} & {\color{blue}\textbf{0.984}}   & {\color{purple}\textit{0.929}} & {\color{purple}\textit{0.956}} & {\color{purple}\textit{0.956}} & {\color{purple}\textit{0.963}} \\
															  & median & 0.631                          & {\color{purple}\textit{1.000}} & 0.999                          & 0.996                          & 0.995                          & 0.998                          & {\color{purple}\textit{0.999}} & {\color{purple}\textit{0.999}} & 0.991                          & {\color{purple}\textit{1.000}} & {\color{blue}\textbf{1.000}}   & 1.000                          & {\color{purple}\textit{1.000}} \\
		\multirow{2}*{Beale\textsuperscript{\dag}}            & mean   & {\color{purple}\textit{0.894}} & {\color{purple}\textit{0.876}} & {\color{purple}\textit{0.876}} & 0.843                          & 0.818                          & {\color{purple}\textit{0.871}} & 0.826                          & {\color{purple}\textit{0.877}} & 0.639                          & {\color{purple}\textit{0.897}} & {\color{blue}\textbf{0.914}}   & {\color{purple}\textit{0.898}} & {\color{purple}\textit{0.890}} \\
															  & median & {\color{purple}\textit{0.984}} & {\color{purple}\textit{0.965}} & {\color{purple}\textit{0.969}} & {\color{purple}\textit{0.943}} & 0.946                          & {\color{purple}\textit{0.941}} & {\color{purple}\textit{0.963}} & {\color{purple}\textit{0.982}} & 0.814                          & {\color{blue}\textbf{0.988}}   & {\color{purple}\textit{0.984}} & {\color{purple}\textit{0.976}} & {\color{purple}\textit{0.979}} \\
		\multirow{2}*{Bohachevsky}                            & mean   & 0.959                          & 0.995                          & 0.995                          & 0.970                          & 0.966                          & 0.996                          & 0.997                          & 0.992                          & 0.994                          & 0.998                          & 0.999                          & {\color{blue}\textbf{1.000}}   & 1.000                          \\
															  & median & 0.998                          & 0.998                          & 0.998                          & 0.996                          & 0.994                          & 0.998                          & 0.999                          & 0.998                          & 0.999                          & 1.000                          & 1.000                          & {\color{blue}\textbf{1.000}}   & 1.000                          \\
		\multirow{2}*{Brochu (2d)\textsuperscript{\dag}}      & mean   & {\color{purple}\textit{0.757}} & 0.772                          & 0.807                          & 0.866                          & 0.902                          & 0.744                          & 0.803                          & 0.751                          & {\color{blue}\textbf{0.932}}   & {\color{purple}\textit{0.818}} & {\color{purple}\textit{0.843}} & {\color{purple}\textit{0.891}} & {\color{purple}\textit{0.929}} \\
															  & median & 0.860                          & 0.871                          & 0.941                          & {\color{purple}\textit{0.969}} & {\color{purple}\textit{0.967}} & 0.792                          & 0.877                          & 0.900                          & {\color{purple}\textit{0.989}} & {\color{purple}\textit{1.000}} & 1.000                          & {\color{blue}\textbf{1.000}}   & {\color{purple}\textit{1.000}} \\
		\multirow{2}*{Brochu (4d)\textsuperscript{\dag}}      & mean   & 0.688                          & 0.628                          & 0.718                          & {\color{blue}\textbf{0.825}}   & {\color{purple}\textit{0.786}} & 0.659                          & 0.702                          & {\color{purple}\textit{0.806}} & 0.254                          & {\color{purple}\textit{0.720}} & {\color{purple}\textit{0.790}} & {\color{purple}\textit{0.748}} & {\color{purple}\textit{0.765}} \\
															  & median & 0.759                          & 0.679                          & {\color{purple}\textit{0.775}} & {\color{purple}\textit{0.863}} & {\color{purple}\textit{0.805}} & 0.719                          & 0.740                          & {\color{purple}\textit{0.890}} & 0.308                          & {\color{purple}\textit{0.771}} & {\color{blue}\textbf{0.923}}   & {\color{purple}\textit{0.793}} & {\color{purple}\textit{0.814}} \\
		\multirow{2}*{Brochu (6d)}                            & mean   & 0.159                          & 0.334                          & 0.336                          & 0.317                          & 0.377                          & 0.343                          & 0.467                          & 0.517                          & 0.106                          & {\color{blue}\textbf{0.714}}   & {\color{purple}\textit{0.702}} & {\color{purple}\textit{0.669}} & {\color{purple}\textit{0.698}} \\
															  & median & 0.131                          & 0.281                          & 0.305                          & 0.306                          & 0.355                          & 0.321                          & 0.396                          & 0.498                          & 0.000                          & {\color{blue}\textbf{0.753}}   & {\color{purple}\textit{0.731}} & {\color{purple}\textit{0.696}} & {\color{purple}\textit{0.727}} \\
		\multirow{2}*{Bukin}                                  & mean   & 0.829                          & {\color{purple}\textit{0.938}} & {\color{purple}\textit{0.907}} & 0.876                          & 0.815                          & {\color{purple}\textit{0.933}} & {\color{purple}\textit{0.926}} & {\color{blue}\textbf{0.959}}   & 0.883                          & 0.882                          & 0.874                          & 0.796                          & 0.747                          \\
															  & median & 0.963                          & {\color{purple}\textit{0.979}} & {\color{purple}\textit{0.973}} & {\color{purple}\textit{0.940}} & 0.891                          & {\color{purple}\textit{0.975}} & {\color{blue}\textbf{0.980}}   & {\color{purple}\textit{0.973}} & 0.912                          & 0.913                          & 0.910                          & 0.924                          & 0.862                          \\
		\multirow{2}*{Camel hump (3d)\textsuperscript{\dag}}  & mean   & {\color{purple}\textit{0.947}} & 0.862                          & {\color{purple}\textit{0.918}} & 0.893                          & 0.920                          & {\color{purple}\textit{0.942}} & 0.890                          & 0.868                          & 0.815                          & {\color{purple}\textit{0.950}} & {\color{purple}\textit{0.968}} & {\color{purple}\textit{0.965}} & {\color{blue}\textbf{0.970}}   \\
															  & median & {\color{purple}\textit{0.989}} & 0.974                          & 0.980                          & 0.973                          & 0.978                          & 0.977                          & 0.985                          & 0.962                          & 0.951                          & 0.991                          & 0.986                          & {\color{blue}\textbf{0.998}}   & {\color{purple}\textit{0.994}} \\
		\multirow{2}*{Camel hump (6d)}                        & mean   & {\color{blue}\textbf{0.983}}   & 0.954                          & 0.949                          & 0.941                          & 0.948                          & 0.948                          & 0.924                          & 0.934                          & 0.870                          & {\color{purple}\textit{0.907}} & {\color{purple}\textit{0.923}} & {\color{purple}\textit{0.916}} & {\color{purple}\textit{0.909}} \\
															  & median & {\color{purple}\textit{0.995}} & {\color{purple}\textit{0.988}} & {\color{purple}\textit{0.982}} & 0.962                          & {\color{purple}\textit{0.977}} & {\color{purple}\textit{0.979}} & {\color{purple}\textit{0.966}} & {\color{purple}\textit{0.983}} & 0.914                          & {\color{blue}\textbf{0.999}}   & {\color{purple}\textit{0.998}} & {\color{purple}\textit{0.997}} & {\color{purple}\textit{0.998}} \\
		\multirow{2}*{Dixon-Price (2d)}                       & mean   & 0.920                          & 0.937                          & 0.945                          & 0.953                          & 0.932                          & 0.941                          & 0.944                          & 0.902                          & 0.860                          & {\color{purple}\textit{0.990}} & {\color{purple}\textit{0.994}} & {\color{purple}\textit{0.994}} & {\color{blue}\textbf{0.995}}   \\
															  & median & 0.972                          & 0.989                          & 0.990                          & 0.980                          & 0.992                          & 0.994                          & 0.988                          & 0.987                          & 0.992                          & {\color{blue}\textbf{0.999}}   & {\color{purple}\textit{0.998}} & {\color{purple}\textit{0.998}} & {\color{purple}\textit{0.998}} \\
		\multirow{2}*{Dixon-Price (4d)\textsuperscript{\dag}} & mean   & {\color{purple}\textit{0.961}} & {\color{purple}\textit{0.981}} & {\color{purple}\textit{0.979}} & {\color{purple}\textit{0.967}} & {\color{purple}\textit{0.968}} & {\color{blue}\textbf{0.983}}   & {\color{purple}\textit{0.982}} & {\color{purple}\textit{0.956}} & 0.955                          & 0.935                          & {\color{purple}\textit{0.935}} & {\color{purple}\textit{0.905}} & {\color{purple}\textit{0.948}} \\
															  & median & {\color{purple}\textit{0.996}} & {\color{blue}\textbf{0.998}}   & {\color{purple}\textit{0.998}} & {\color{purple}\textit{0.998}} & {\color{purple}\textit{0.996}} & {\color{purple}\textit{0.998}} & {\color{purple}\textit{0.998}} & {\color{purple}\textit{0.995}} & 0.983                          & 0.989                          & {\color{purple}\textit{0.991}} & {\color{purple}\textit{0.995}} & {\color{purple}\textit{0.995}} \\
		\multirow{2}*{Drop-wave\textsuperscript{\dag}}        & mean   & {\color{purple}\textit{0.691}} & {\color{purple}\textit{0.770}} & {\color{purple}\textit{0.782}} & {\color{purple}\textit{0.787}} & {\color{blue}\textbf{0.820}}   & 0.762                          & {\color{purple}\textit{0.777}} & {\color{purple}\textit{0.757}} & 0.530                          & 0.650                          & 0.694                          & 0.687                          & 0.630                          \\
															  & median & {\color{purple}\textit{0.789}} & {\color{purple}\textit{0.834}} & {\color{purple}\textit{0.883}} & {\color{purple}\textit{0.903}} & {\color{blue}\textbf{0.907}}   & 0.827                          & {\color{purple}\textit{0.883}} & {\color{purple}\textit{0.840}} & 0.573                          & 0.733                          & 0.832                          & 0.832                          & 0.717                          \\
		\multirow{2}*{Egg holder}                             & mean   & 0.461                          & {\color{purple}\textit{0.636}} & {\color{purple}\textit{0.614}} & 0.599                          & 0.604                          & 0.608                          & {\color{purple}\textit{0.634}} & 0.586                          & {\color{blue}\textbf{0.721}}   & 0.480                          & 0.503                          & 0.521                          & 0.468                          \\
															  & median & 0.497                          & {\color{purple}\textit{0.653}} & {\color{purple}\textit{0.652}} & 0.612                          & 0.641                          & 0.643                          & {\color{purple}\textit{0.672}} & 0.639                          & {\color{blue}\textbf{0.762}}   & 0.461                          & 0.534                          & 0.516                          & 0.463                          \\
		\multirow{2}*{Griewank}                               & mean   & 0.736                          & {\color{purple}\textit{0.921}} & 0.881                          & 0.881                          & 0.805                          & {\color{purple}\textit{0.968}} & {\color{blue}\textbf{0.971}}   & {\color{purple}\textit{0.969}} & {\color{purple}\textit{0.969}} & {\color{purple}\textit{0.966}} & {\color{purple}\textit{0.962}} & {\color{purple}\textit{0.964}} & {\color{purple}\textit{0.967}} \\
															  & median & 0.764                          & {\color{purple}\textit{0.975}} & 0.964                          & 0.943                          & 0.902                          & {\color{purple}\textit{0.975}} & {\color{purple}\textit{0.979}} & {\color{purple}\textit{0.978}} & {\color{purple}\textit{0.979}} & {\color{blue}\textbf{0.981}}   & {\color{purple}\textit{0.979}} & {\color{purple}\textit{0.975}} & {\color{purple}\textit{0.976}} \\
		\multirow{2}*{Hartman (3d)}                           & mean   & 0.608                          & 0.754                          & 0.781                          & 0.767                          & 0.804                          & 0.847                          & 0.915                          & 0.894                          & 0.917                          & {\color{blue}\textbf{0.984}}   & {\color{purple}\textit{0.982}} & 0.984                          & 0.984                          \\
															  & median & 0.579                          & 0.817                          & 0.867                          & 0.829                          & 0.854                          & 0.961                          & 0.969                          & 0.982                          & 0.944                          & {\color{blue}\textbf{1.000}}   & {\color{purple}\textit{1.000}} & 1.000                          & 1.000                          \\
		\multirow{2}*{Hartman (6d)\textsuperscript{\dag}}     & mean   & {\color{purple}\textit{0.961}} & 0.889                          & 0.897                          & 0.943                          & 0.906                          & 0.909                          & 0.917                          & 0.928                          & 0.889                          & {\color{purple}\textit{0.969}} & {\color{purple}\textit{0.958}} & {\color{blue}\textbf{0.973}}   & 0.931                          \\
															  & median & {\color{blue}\textbf{0.999}}   & 0.909                          & 0.936                          & 0.964                          & 0.942                          & 0.957                          & 0.947                          & 0.949                          & 0.915                          & {\color{purple}\textit{0.998}} & {\color{purple}\textit{0.996}} & {\color{purple}\textit{0.997}} & {\color{purple}\textit{0.997}} \\
		\multirow{2}*{Himmelblau}                             & mean   & 0.968                          & 0.973                          & 0.968                          & 0.980                          & 0.970                          & 0.964                          & 0.973                          & 0.919                          & 0.978                          & {\color{purple}\textit{0.996}} & {\color{blue}\textbf{0.996}}   & {\color{purple}\textit{0.994}} & {\color{purple}\textit{0.995}} \\
															  & median & 0.997                          & 0.992                          & 0.992                          & 0.988                          & 0.987                          & 0.994                          & 0.991                          & 0.995                          & 0.995                          & {\color{purple}\textit{0.999}} & {\color{blue}\textbf{1.000}}   & {\color{purple}\textit{0.999}} & {\color{purple}\textit{0.998}} \\
		\multirow{2}*{Levy (2d)}                              & mean   & {\color{purple}\textit{0.914}} & {\color{purple}\textit{0.940}} & {\color{purple}\textit{0.937}} & 0.870                          & 0.866                          & {\color{blue}\textbf{0.941}}   & {\color{purple}\textit{0.934}} & {\color{purple}\textit{0.940}} & 0.763                          & {\color{purple}\textit{0.919}} & 0.883                          & 0.797                          & 0.841                          \\
															  & median & {\color{purple}\textit{0.990}} & {\color{purple}\textit{0.988}} & {\color{purple}\textit{0.984}} & 0.972                          & 0.951                          & {\color{blue}\textbf{0.993}}   & {\color{purple}\textit{0.981}} & {\color{purple}\textit{0.984}} & 0.875                          & {\color{purple}\textit{0.993}} & 0.968                          & 0.875                          & 0.966                          \\
		\multirow{2}*{Levy (4d)\textsuperscript{\dag}}        & mean   & {\color{blue}\textbf{0.968}}   & 0.939                          & 0.910                          & 0.915                          & 0.909                          & 0.951                          & 0.933                          & 0.938                          & 0.758                          & 0.868                          & 0.865                          & 0.839                          & 0.893                          \\
															  & median & {\color{blue}\textbf{0.983}}   & 0.960                          & 0.956                          & 0.933                          & 0.956                          & 0.970                          & 0.946                          & 0.951                          & 0.814                          & 0.947                          & 0.923                          & 0.906                          & 0.927                          \\
		\multirow{2}*{Levy (6d)\textsuperscript{\dag}}        & mean   & {\color{purple}\textit{0.954}} & 0.931                          & 0.913                          & 0.933                          & 0.926                          & {\color{blue}\textbf{0.964}}   & 0.942                          & 0.908                          & 0.850                          & 0.874                          & 0.860                          & 0.890                          & 0.857                          \\
															  & median & {\color{purple}\textit{0.972}} & 0.957                          & 0.942                          & 0.948                          & 0.958                          & {\color{blue}\textbf{0.976}}   & 0.962                          & 0.944                          & 0.860                          &  0.871                         & 0.908                          & 0.931                          & 0.849                          \\
		\multirow{2}*{Powell}                                 & mean   & {\color{blue}\textbf{0.957}}   & 0.942                          & 0.918                          & 0.911                          & 0.853                          & 0.947                          & 0.903                          & {\color{purple}\textit{0.944}} & 0.884                          & 0.923                          & 0.903                          & {\color{purple}\textit{0.908}} & 0.896                          \\
															  & median & {\color{purple}\textit{0.986}} & {\color{purple}\textit{0.975}} & {\color{purple}\textit{0.982}} & {\color{purple}\textit{0.973}} & 0.941                          & {\color{purple}\textit{0.969}} & {\color{purple}\textit{0.975}} & {\color{purple}\textit{0.981}} & 0.943                          & {\color{purple}\textit{0.982}} & {\color{purple}\textit{0.980}} & {\color{blue}\textbf{0.990}}   & {\color{purple}\textit{0.979}} \\
		\multirow{2}*{Rastrigin\textsuperscript{\dag}}        & mean   & 0.627                          & 0.672                          & 0.623                          & 0.625                          & 0.615                          & {\color{purple}\textit{0.754}} & 0.743                          & 0.735                          & 0.718                          & {\color{purple}\textit{0.785}} & {\color{blue}\textbf{0.800}}   & 0.742                          & {\color{purple}\textit{0.764}} \\
															  & median & 0.630                          & 0.689                          & 0.645                          & 0.592                          & 0.599                          & {\color{purple}\textit{0.810}} & 0.767                          & 0.764                          & 0.750                          & {\color{purple}\textit{0.801}} & {\color{blue}\textbf{0.850}}   & 0.789                          & {\color{purple}\textit{0.781}} \\
		\multirow{2}*{Rosenbrock\textsuperscript{\dag}}       & mean   & {\color{purple}\textit{0.930}} & {\color{purple}\textit{0.919}} & {\color{blue}\textbf{0.961}}   & {\color{purple}\textit{0.948}} & {\color{purple}\textit{0.924}} & {\color{purple}\textit{0.927}} & {\color{purple}\textit{0.957}} & {\color{purple}\textit{0.959}} & 0.935                          & 0.914                          & 0.928                          & 0.905                          & 0.864                          \\
															  & median & {\color{purple}\textit{0.984}} & {\color{purple}\textit{0.977}} & {\color{purple}\textit{0.982}} & {\color{purple}\textit{0.977}} & {\color{purple}\textit{0.983}} & 0.978                          & {\color{blue}\textbf{0.986}}   & {\color{purple}\textit{0.980}} & 0.974                          & 0.964                          & 0.957                          & {\color{purple}\textit{0.982}} & 0.964                          \\
		\multirow{2}*{Shubert}                                & mean   & 0.347                          & {\color{purple}\textit{0.512}} & {\color{purple}\textit{0.469}} & {\color{purple}\textit{0.504}} & {\color{blue}\textbf{0.555}}   & {\color{purple}\textit{0.483}} & {\color{purple}\textit{0.487}} & 0.375                          & 0.223                          & 0.361                          & {\color{purple}\textit{0.388}} & 0.370                          & 0.387                          \\
															  & median & 0.222                          & {\color{purple}\textit{0.416}} & {\color{purple}\textit{0.359}} & {\color{purple}\textit{0.390}} & {\color{blue}\textbf{0.540}}   & {\color{purple}\textit{0.390}} & {\color{purple}\textit{0.403}} & 0.236                          & 0.160                          & 0.198                          & {\color{purple}\textit{0.234}} & 0.273                          & 0.233                          \\
		\multirow{2}*{Step 2\textsuperscript{\dag}}           & mean   & {\color{blue}\textbf{1.000}}   & 0.991                          & 0.991                          & 0.987                          & 0.987                          & 0.998                          & 0.995                          & 0.995                          & 0.990                          & 0.998                          & {\color{purple}\textit{0.999}} & 0.999                          & 0.999                          \\
															  & median & {\color{purple}\textit{1.000}} & 0.997                          & 0.993                          & 0.985                          & 0.988                          & 0.999                          & 0.997                          & 0.997                          & 0.993                          & 0.999                          & {\color{blue}\textbf{1.000}}   & 0.999                          & {\color{purple}\textit{0.999}} \\
		\multirow{2}*{Styblinski-Tang}                        & mean   & 0.483                          & 0.597                          & 0.611                          & 0.566                          & 0.482                          & 0.663                          & 0.662                          & {\color{purple}\textit{0.652}} & 0.000                          & {\color{blue}\textbf{0.735}}   & {\color{purple}\textit{0.706}} & {\color{purple}\textit{0.702}} & {\color{purple}\textit{0.724}} \\
															  & median & 0.456                          & 0.577                          & 0.634                          & 0.586                          & 0.476                          & 0.688                          & {\color{purple}\textit{0.691}} & {\color{purple}\textit{0.657}} & 0.000                          & {\color{purple}\textit{0.739}} & {\color{purple}\textit{0.701}} & {\color{purple}\textit{0.729}} & {\color{blue}\textbf{0.749}}   \\ 
		\midrule
		\multirow{2}*{Cosmological\textsuperscript{\dag}}     & mean   & 0.959                          & 0.960                          & 0.960                          & 0.956                          & 0.955                          & 0.962                          & 0.959                          & 0.936                          & 0.897                          & {\color{purple}\textit{0.960}} & {\color{purple}\textit{0.953}} & {\color{blue}\textbf{0.968}}   & {\color{purple}\textit{0.961}} \\
															  & median & 0.971                          & 0.972                          & 0.973                          & 0.971                          & 0.970                          & 0.975                          & 0.971                          & 0.968                          & 0.952                          & {\color{purple}\textit{0.974}} & {\color{purple}\textit{0.977}} & {\color{blue}\textbf{0.978}}   & {\color{purple}\textit{0.975}} \\
		\multirow{2}*{LDA}                                    & mean   & 0.752                          & {\color{purple}\textit{0.836}} & 0.796                          & 0.759                          & 0.718                          & {\color{purple}\textit{0.880}} & {\color{purple}\textit{0.856}} & {\color{blue}\textbf{0.901}}   & 0.316                          & 0.669                          & 0.624                          & 0.658                          & 0.689                          \\
															  & median & 0.886                          & {\color{purple}\textit{0.906}} & 0.837                          & 0.803                          & 0.825                          & {\color{purple}\textit{0.921}} & {\color{purple}\textit{0.909}} & {\color{blue}\textbf{0.939}}   & 0.281                          & 0.771                          & 0.740                          & 0.685                          & 0.790                          \\
		\multirow{2}*{LogReg}                                 & mean   & 0.464                          & 0.732                          & 0.640                          & 0.602                          & 0.589                          & 0.644                          & {\color{blue}\textbf{0.858}}   & {\color{purple}\textit{0.850}} & 0.559                          & {\color{purple}\textit{0.854}} & {\color{purple}\textit{0.851}} & {\color{purple}\textit{0.828}} & {\color{purple}\textit{0.845}} \\
															  & median & 0.352                          & 0.827                          & 0.766                          & 0.788                          & 0.739                          & 0.847                          & {\color{purple}\textit{0.966}} & {\color{purple}\textit{0.960}} & 0.682                          & {\color{blue}\textbf{0.981}}   & {\color{purple}\textit{0.977}} & {\color{purple}\textit{0.977}} & {\color{purple}\textit{0.966}} \\
		\multirow{2}*{NN Boston\textsuperscript{\dag}}        & mean   & 0.274                          & {\color{purple}\textit{0.406}} & 0.360                          & {\color{purple}\textit{0.381}} & {\color{purple}\textit{0.422}} & {\color{purple}\textit{0.362}} & {\color{purple}\textit{0.379}} & {\color{blue}\textbf{0.436}}   & 0.326                          & 0.366                          & {\color{purple}\textit{0.394}} & {\color{purple}\textit{0.420}} & {\color{purple}\textit{0.431}} \\
															  & median & 0.240                          & {\color{purple}\textit{0.467}} & {\color{purple}\textit{0.380}} & 0.415                          & {\color{purple}\textit{0.483}} & {\color{purple}\textit{0.408}} & {\color{purple}\textit{0.392}} & {\color{purple}\textit{0.458}} & 0.375                          & {\color{purple}\textit{0.355}} & {\color{purple}\textit{0.436}} & {\color{purple}\textit{0.479}} & {\color{blue}\textbf{0.497}}   \\
		\multirow{2}*{NN Cancer\textsuperscript{\dag}}        & mean   & {\color{purple}\textit{0.609}} & {\color{purple}\textit{0.617}} & {\color{purple}\textit{0.645}} & {\color{purple}\textit{0.629}} & {\color{purple}\textit{0.622}} & {\color{purple}\textit{0.669}} & {\color{purple}\textit{0.628}} & {\color{blue}\textbf{0.687}}   & 0.498                          & 0.542                          & {\color{purple}\textit{0.570}} & 0.539                          & 0.468                          \\
															  & median & {\color{purple}\textit{0.702}} & {\color{purple}\textit{0.702}} & {\color{purple}\textit{0.730}} & {\color{purple}\textit{0.705}} & {\color{purple}\textit{0.702}} & {\color{purple}\textit{0.730}} & {\color{purple}\textit{0.705}} & {\color{blue}\textbf{0.782}}   & 0.609                          & 0.670                          & {\color{purple}\textit{0.654}} & 0.615                          & 0.609                          \\
		\multirow{2}*{Robot push (3d)}                        & mean   & 0.584                          & 0.779                          & 0.704                          & 0.664                          & 0.612                          & {\color{purple}\textit{0.834}} & {\color{purple}\textit{0.846}} & {\color{blue}\textbf{0.859}}   & 0.726                          & 0.664                          & 0.642                          & 0.597                          & 0.702                          \\
															  & median & 0.593                          & 0.838                          & 0.790                          & 0.707                          & 0.719                          & {\color{purple}\textit{0.889}} & {\color{purple}\textit{0.893}} & {\color{blue}\textbf{0.904}}   & 0.818                          & 0.829                          & 0.813                          & 0.747                          & 0.839                          \\
		\multirow{2}*{Robot push (4d)}                        & mean   & 0.317                          & 0.464                          & 0.411                          & 0.428                          & 0.399                          & 0.534                          & {\color{purple}\textit{0.582}} & {\color{blue}\textbf{0.618}}   & 0.412                          & 0.460                          & 0.410                          & 0.379                          & 0.399                          \\
															  & median & 0.319                          & 0.488                          & 0.436                          & 0.464                          & 0.384                          & 0.559                          & {\color{purple}\textit{0.574}} & {\color{blue}\textbf{0.645}}   & 0.463                          & 0.502                          & 0.446                          & 0.401                          & 0.447                          \\
		\multirow{2}*{SVM}                                    & mean   & {\color{purple}\textit{0.927}} & {\color{purple}\textit{0.952}} & {\color{purple}\textit{0.943}} & 0.911                          & 0.918                          & {\color{purple}\textit{0.946}} & {\color{purple}\textit{0.952}} & {\color{blue}\textbf{0.965}}   & 0.613                          & 0.780                          & 0.775                          & 0.768                          & 0.842                          \\
															  & median & {\color{purple}\textit{0.980}} & {\color{purple}\textit{0.984}} & {\color{purple}\textit{0.978}} & 0.977                          & 0.973                          & {\color{purple}\textit{0.981}} & {\color{purple}\textit{0.983}} & {\color{blue}\textbf{0.986}}   & 0.791                          & 0.959                          & 0.947                          & 0.945                          & 0.954                          \\
	\end{longtable}
\end{landscape}
}  

The results shown in \Cref{tab:numerical:synthetic-and-real:gap} and \Cref{fig:numerical:itertime-vs-gap} corroborate the findings of the other related works. 
In line with \cite{bemporad_global_2020,bemporad_active_2023}, non-Bayesian methods retain comparable or higher performance compared to Bayesian ones, but exhibit lower computation times. Myopic method \textit{B} is $55.94\% \pm 7.69\%$ faster than \textit{EI}, whereas we see on average a reduction of $43.18\% \pm 4.86\%$ from nonmyopic methods \textit{EIR} to \textit{R}. One can notice that in general nonmyopic combinations outperform their myopic counterparts, with the latter achieving statistically comparable or better results only for a small subset of the problems. This empirically validates the efficacy of exploiting a lookahead strategy to overcome the drawbacks of single-state acquisition functions. Note that, however, longer rollout horizons are not necessarily associated with better optimality gaps. This observation is in accordance with the rest of the BO and control state of the art \citep{yue_why_2020,shi_suboptimality_2023}: while predicting further into the future is generally desirable, it may be counterproductive due to model mismatches; in practice, shorter horizons (e.g., 2 and 3) may be the most beneficial in striking a good balance between performance improvement and computational demands.
On the other hand, multi-step approaches tend to gain more performance as the horizon increases, often ranking as the most performing method (or significantly close to it), especially for the real-world benchmarks. This result validates the intuition that the multi-step branching formulation is more likely to capture the real unknown dynamics of the surrogate model and to suffer less from model error propagation, but it comes at the price of a higher computational burden.

We also report in \Cref{fig:numerical:gaps-benchmarks-synthetic,fig:numerical:gaps-benchmarks-real} the evolution of the optimality gap during the sequential optimisation iterations. For clarity, here we include only the base myopic strategy and the best nonmyopic one among the proposed. It can be seen that, in most cases, not only does the nonmyopic algorithm converge to a better gap, but it also does so often in a similar or faster rate than its myopic counterpart \textit{B}. This property is especially desirable in contexts in which the budget, e.g., the maximum number of iterations $N$, is limited \citep{lam_bayesian_2016,lee_nonmyopic_2021}.
\begin{figure}
    \centering
    \input{imgs/gaps-benchmarks-synthetic}
    \caption{Comparison of the evolution of the optimality gap during optimisation (excluding the $N_0$ warm-up iterations) of each synthetic benchmark problem for the myopic method \textit{B} (dashed) and the best mean-performing proposed nonmyopic method (solid), as reported in \Cref{tab:numerical:synthetic-and-real:gap}. The colour scheme follows from \Cref{fig:numerical:itertime-vs-gap}. Averages and $95\%$ confidence intervals are computed over the 30 repetitions, and iterations have been normalised to the range $[0,1]$. An optimality gap of 1 indicates convergence to the optimal solution}
    \label{fig:numerical:gaps-benchmarks-synthetic}
\end{figure}
\begin{figure}
    \centering
    \input{imgs/gaps-benchmarks-real}
    \caption{Comparison of the evolution of the optimality gap during optimisation (excluding the $N_0$ warm-up iterations) of each real-world benchmark problem for the myopic method \textit{B} (dashed) and the best mean-performing proposed nonmyopic method (solid), as reported in \Cref{tab:numerical:synthetic-and-real:gap}. The colour scheme follows from \Cref{fig:numerical:itertime-vs-gap}. Averages and $95\%$ confidence intervals are computed over the 30 repetitions, and iterations have been normalised to the range $[0,1]$. An optimality gap of 1 indicates convergence to the optimal solution}
    \label{fig:numerical:gaps-benchmarks-real}
\end{figure}

\subsubsection{Example with Constrained Search Space}

To showcase the ability of the proposed methodology in handling problems with nonconvex nonlinear constraints, we also test the proposed rollout strategy \textit{R-2} on an additional synthetic problem, the Gramacy constrained test function \citep{gramacy_modeling_2016}. The problem features two nonlinear nonconvex constraints: 
\begin{mini}
    { \scriptstyle 0 \leq x_1, x_2 \leq 1 }{ x_1 + x_2 }{}{}
    \addConstraint{ x_1 + 2 x_2 + \frac{1}{2} \sin \left( 2 \pi \left( x_1^2 - 2 x_2 \right) \right) \geq \frac{3}{2}, }
    \addConstraint{ x_1^2 + x_2^2 \leq \frac{3}{2}. }
\end{mini}
To warm the surrogate model up only on feasible locations, the $N_0$ initial points are drawn via constrained Latin hypercube sampling rather than uniformly at random  \citep[see][Algorithm~2]{bemporad_global_2020}. SLSQP \citep{kraft_algorithm_1994} is employed to solve the nonlinear constrained acquisition problem instead of L-BFGS-B, since the latter cannot handle such kind of optimisation problems. The rest of the rollout algorithm's hyperparameters are selected as explained in the previous section. \Cref{fig:numerical:gramacy} reports the evolution of queried locations in the search space as the algorithm converges. It can be clearly noticed that the constraints are effectively taken into consideration, with all query points lying inside the feasible region defined by the two constraints and the box bounds.
\begin{figure}
    \centering
    \input{imgs/gramacy}
    \caption{Evolution of the points queried by \textit{R-2} in search for the optimal location (red star) of the Gramacy constrained test function. The greyed areas indicate the infeasible portions of the search space, while the straight lines inside the feasible region are the level curves of the objective function}
    \label{fig:numerical:gramacy}
\end{figure}

\subsection{Data-driven Tuning of an MPC Controller}

The second numerical experiment consists of utilising the proposed methodology to automatically tune an MPC controller for a continuously stirred tank reactor system. This system, adapted from \cite{subramanian_handling_2015,sorourifar_data_2021}, undergoes the set of reactions
\begin{equation}
    \text{A} \rightarrow \text{B} \rightarrow \text{C}, \quad 2 \text{A} \rightarrow \text{D}.
\end{equation}
where A, B, C and D identify different chemical reagents. Its nonlinear dynamics are given by the following differential equations:
\begin{align}
    \dot{c}_\text{A} ={}& F \left( c_{\text{A},0} - c_\text{A} \right) - k_1 c_\text{A} - k_3 c_\text{A}^2, \label{eq:numerical:cstr:dynamics:first}  \\
    \dot{c}_\text{B} ={}& - F c_\text{B} + k_1 c_\text{A} - k_2 c_\text{B}, \\
    \begin{split}
        \dot{T}_\text{R} ={}& F \left( T_\text{in} - T_\text{R} \right)
            + \frac{ k_\text{W} \text{A} }{ \rho c_\text{p} V_\text{R} } \left( T_\text{K} - T_\text{R} \right) \\
            &- \frac{ k_1 c_\text{A} \Delta H_\text{AB} + k_2 c_\text{B} \Delta H_\text{BC} + k_3 c_\text{A}^2 \Delta H_\text{AD} }{ \rho c_\text{p} },
    \end{split} \\
    \dot{T}_\text{K} ={}& \frac{ \dot{Q}_\text{K} + k_\text{W} A \left( T_\text{R} - T_\text{K} \right) }{ m_\text{K} c_\text{pK} }, \label{eq:numerical:cstr:dynamics:last}
\end{align}
with
\begin{equation}
    k_i = k_{0,i} \exp{ \left( - \frac{ E_{A,i} }{ R \left( T_\text{R} + 273.15 \right) } \right) }, \quad i = 1,2,3,
\end{equation}
where the feed flowrate $F \in \mathbb{R}$ is the control input, and the concentrations $c_\text{A}, c_\text{B} \in \mathbb{R}$ of the corresponding components and the reactor and coolant temperatures $T_\text{R},T_\text{K} \in \mathbb{R}$ form the state (please refer to \cite{subramanian_handling_2015} for more details on the dynamics and its coefficients). At any sampled time $t$, the whole state is unknown and only the concentration of species B and the reactor temperature are assumed to be measurable with some additive noise, so that
\begin{equation}
    y_t = \begin{bmatrix} c_{\text{B}} \\ T_{\text{R}} \end{bmatrix} + w_t, \quad w_t \sim \mathcal{N}\left(
        \begin{bmatrix} 0 \\ 0 \end{bmatrix}, 
        \begin{bmatrix} 0.2^2 & 0 \\ 0 & 10^2 \end{bmatrix}
    \right).
\end{equation}
The goal is to control the system in such a way as to maximise the production of component B, expressed by the instantaneous production of moles of this component $\dot{n}_\text{B} = V_\text{R} c_\text{B} F$. At the same time, the controller should strive to constrain the temperature of the reactor within the range $[100, 150] \ \unit{\degreeCelsius}$. Therefore, we discretise time with the sampling interval $t_0 = \qty{0.005}{\hour}$ and deploy a piecewise-constant control policy. At sampled time $t$, this policy is provided by the following parametric discrete-time MPC controller with horizon $h_\text{c} = 10$:
\begin{mini!}
    {
        \scriptstyle
        \substack{
            \hat{y}_{t|t},\dots,\hat{y}_{t+h_\text{c}|t}\\
            \widehat{F}_{t|t},\dots,\widehat{F}_{t+h_\text{c}-1|t}\\
            \underline\sigma_t,\dots,\underline\sigma_{t+h_\text{c}}\\
            \overline\sigma_t,\dots,\overline\sigma_{t+h_\text{c}}
        }
    }{
        \sum_{j=t}^{t+h_\text{c}}{ w \left( \underline\sigma_j + \overline\sigma_j \right) }
        - \sum_{j=t}^{t+h_\text{c}-1}{ V_\text{R} \hat{c}_{\text{B},j|t} \widehat{F}_{j|t} }
        \label{eq:numerical:cstr:mpc:objective}
    }{ \label{eq:numerical:cstr:mpc} }{}
    \addConstraint{ \hat{y}_{t|t} = y_t, }
    \addConstraint{
        \hat{y}_{j+1|t} = \mathcal{F}_{\theta_\text{m}} \left( \hat{y}_{j|t}, \widehat{F}_{j|t} \right), \quad
        \label{eq:numerical:cstr:mpc:dynamics}
    }{}{ j = t,\dots,t+h_\text{c}-1, }
    \addConstraint{
         5 \le \widehat{F}_{j|t} \le 35,
    }{}{ j = t,\dots,t+h_\text{c}-1, }
    \addConstraint{
         \widehat{T}_{\text{R},j|t} \ge 100 + \theta_\text{b} + \underline\sigma_j,
         \label{eq:numerical:cstr:mpc:temp:lb}
    }{}{ j = t,\dots,t+h_\text{c}, }
    \addConstraint{
         \widehat{T}_{\text{R},j|t} \le 150 - \theta_\text{b} + \overline\sigma_j,
         \label{eq:numerical:cstr:mpc:temp:ub}
    }{}{ j = t,\dots,t+h_\text{c}, }
    \addConstraint{
         \overline\sigma_j \ge 0, \ \underline\sigma_j \ge 0,
    }{}{ j = t,\dots,t+h_\text{c}. }
\end{mini!}
The constraints on the temperature \eqref{eq:numerical:cstr:mpc:temp:lb}-\eqref{eq:numerical:cstr:mpc:temp:ub} are made soft via the use of the slack variables $\underline\sigma_t$, $\overline\sigma_t$, which are appropriately penalised in the objective \eqref{eq:numerical:cstr:mpc:objective} with a sufficiently high weight $w = 4 \cdot 10^3$. In this way, the controller is encouraged to satisfy the constraints, but the optimisation problem will not become infeasible in case of an unavoidable violation of these. Moreover, a backoff parameter $\theta_\text{b} \in \mathbb{R}$ is included in the temperature constraints. When appropriately tuned, this parameter allows the controller to avoid violations more robustly. As in \cite{sorourifar_data_2021}, we assume the dynamics \eqref{eq:numerical:cstr:dynamics:first}-\eqref{eq:numerical:cstr:dynamics:last} are black-box, i.e., they cannot be used in the proposed model-based control approach. Thus, as prediction model in \eqref{eq:numerical:cstr:mpc:dynamics}, we use the unitary-lag NARX model $\mathcal{F}_{\theta_\text{m}} : \mathbb{R}^2 \times \mathbb{R} \rightarrow \mathbb{R}^2$ defined as
\begin{equation}
        \mathcal{F}_{\theta_\text{m}}\left( y, F \right) = 
        \theta_{\text{m},o} + 
        \begin{bmatrix} \theta_{\text{m},y,1}^\top \\ \theta_{\text{m},y,2}^\top \end{bmatrix} y + 
        \theta_{\text{m},F} F + 
        \begin{bmatrix} \theta_{\text{m},y^2,1}^\top \\ \theta_{\text{m},y^2,2}^\top \end{bmatrix} y^2 + 
        \theta_{\text{m},F^2} F^2,
\end{equation}
with $\theta_\text{m} = \begin{bmatrix} \theta_{\text{m},o}^\top & \theta_{\text{m},y,1}^\top & \theta_{\text{m},y,2}^\top & \theta_{\text{m},F}^\top & \theta_{\text{m},y^2m,1}^\top & \theta_{\text{m},y^2,2}^\top & \theta_{\text{m},F^2}^\top \end{bmatrix}^\top \in \mathbb{R}^{14}$. Alongside $\theta_\text{b}$, these forms the parameter vector $\theta = \begin{bmatrix} \theta_\text{m}^\top & \theta_\text{b} \end{bmatrix}^\top$, totaling to $14+1=15$. As usual, we adopt a receding horizon approach: once \eqref{eq:numerical:cstr:mpc} is solved for $t$, only the first optimal action $\widehat{F}^\star_{t|t}$ is applied to the real system, and at the next time step $t+1$ the procedure is iterated again.

To maximise the production of component B and to minimise constraint violations, a valid numerical value must be assigned to the parameters $\theta$. However, doing so manually can be challenging due to, e.g., the nonlinearities in the system and the high number of parameters to be tuned. Therefore, we propose to exploit the proposed GO strategies to automatically tune these parameters. Consider the objective function
\begin{equation} \label{eq:numerical:mpc-tuning:objective}
    J\left(\theta\right) = \int_{0}^{t_\text{f}}{ \bigl(
        w \max{\{0,100-T_{\text{R}}\}} + 
        w \max{\{0,T_{\text{R}}-150\}} -
        n_{\text{B}}
    \bigr) dt },
\end{equation}
with $t_\text{f} = \qty{0.2}{\hour}$. This function is easy to evaluate via simulation once a fixed value of $\theta$ is given, but is otherwise difficult to formulate or analyse in closed form. Therefore, mirroring \eqref{eq:background:go:problem}, we can pose the tuning problem as the optimisation problem
\begin{argmini}
    { \scriptstyle \theta \in \Theta }{ J(\theta) }{}{ \theta^\star \in },
\end{argmini}
where $\Theta$ bounds the search space to the hyperrectangle $\left[-2, 2\right]^{14} \times \left[0, 10\right]$.

To solve it, we deploy a similar algorithm to \Cref{section:numerical-experiments:benchmarks}. First, a random selection of $N_0=5$ parameters and their corresponding evaluation of the performance of the parametric MPC policy, i.e., $\left\{ \bigl( \theta_i, J(\theta_i) \bigr) \mid i = 1,\dots,5 \right\}$, are used to warm the surrogate model up (in this case, IDW). Then, different GO strategies are given $50$ iterations to seek the optimal parametrisation of the MPC controller. When evaluating $J(\theta_i)$, each episode is limited to $\frac{t_f}{t_0} = 40$ time steps, and the initial conditions are fixed to $c_{\text{A,0}} = c_{\text{B,0}} = \qty{1}{\mole\per\liter}$, $T_{\text{R,0}} = T_{\text{K,0}} = \qty{100}{\degreeCelsius}$ for every simulation. Since the aforementioned NARX model is nonlinear, to solve \eqref{eq:numerical:cstr:mpc} we employ the IPOPT solver \citep{wachter_implementation_2006} alongside a multistart approach with 10 run to mitigate convergence to poor solutions. The evolution of the real reactor dynamics under the MPC control policy is instead computed via numerical integration of \eqref{eq:numerical:cstr:dynamics:first}-\eqref{eq:numerical:cstr:dynamics:last} with CVODES \citep{hindmarsh_sundials_2005}. Since the state observations $y_t$ and the selection of warm-up parameters are stochastic, the experiment is repeated 30 times with different random seeds.
\begin{figure}
    \centering
    \begin{tikzpicture}
    \pgfplotstableread{data/mpc-tuning/gap_cstr-mpc-tuning_myopic.dat}\TuningMyopicData
    \pgfplotstableread{data/mpc-tuning/gap_cstr-mpc-tuning_r4mc.dat}\TuningRFourGhData
    \pgfplotstableread{data/mpc-tuning/gap_cstr-mpc-tuning_ms3mc.dat}\TuningMsThreeGhData

    \pgfplotstableread{data/mpc-tuning/reactor_temp_ms3mc_iter1.dat}\LowIterData
    \pgfplotstableread{data/mpc-tuning/reactor_temp_ms3mc_iter25.dat}\MidIterData
    \pgfplotstableread{data/mpc-tuning/reactor_temp_ms3mc_iter53.dat}\HighIterData
    \def\trials{30}

    \begin{groupplot}[
        group style={
                group name=mpcplots,
                group size=2 by 1,
                horizontal sep=1.7cm,
            },
        height=\axisdefaultheight,
        width=0.5\axisdefaultwidth,
    ]
        \nextgroupplot[
            every axis/.append style={
                const plot mark right,
            },
            xlabel=Iteration,
            ylabel=Optimality gap,
            legend pos=south east,
        ]
            \addplot [C4, densely dashed] table [x expr=\coordindex, y=avg] {\TuningMyopicData};
            \addplot [C4, densely dashed, ultra thin, opacity=0.5, const plot mark right, name path=lb] table [x expr=\coordindex, y=lb] {\TuningMyopicData};
            \addplot [C4, densely dashed, ultra thin, opacity=0.5, const plot mark right, name path=ub] table [x expr=\coordindex, y=ub] {\TuningMyopicData};
            \addplot [C4, densely dashed, opacity=0.2] fill between [of=lb and ub, reverse=true];

            \addplot [C2] table [x expr=\coordindex, y=avg] {\TuningRFourGhData};
            \addplot [C2, ultra thin, opacity=0.5, const plot mark right, name path=lb] table [x expr=\coordindex, y=lb] {\TuningRFourGhData};
            \addplot [C2, ultra thin, opacity=0.5, const plot mark right, name path=ub] table [x expr=\coordindex, y=ub] {\TuningRFourGhData};
            \addplot [C2, opacity=0.2] fill between [of=lb and ub, reverse=true];

            \addplot [C3] table [x expr=\coordindex, y=avg] {\TuningMsThreeGhData};
            \addplot [C3, ultra thin, opacity=0.5, const plot mark right, name path=lb] table [x expr=\coordindex, y=lb] {\TuningMsThreeGhData};
            \addplot [C3, ultra thin, opacity=0.5, const plot mark right, name path=ub] table [x expr=\coordindex, y=ub] {\TuningMsThreeGhData};
            \addplot [C3, opacity=0.2] fill between [of=lb and ub, reverse=true];

            \legend{\textit{B},,,,\textit{R-4},,,,\textit{MS-3}}

        \nextgroupplot[
            xlabel={Time [\unit{\hour}]},
            ylabel={$T_\text{R} \ [\unit{\degreeCelsius}]$},
            every axis/.append style={smooth},
            xticklabel style={
                /pgf/number format/.cd,
                fixed,
            },
        ]
            \addplot [red, semithick, domain=0:0.2, samples=2] {100};

            \addplot [C0, densely dotted] table [x=time, y=avg] {\LowIterData};
            \addplot [C0, densely dotted, ultra thin, opacity=0.5, name path=lb] table [x=time, y=lb] {\LowIterData};
            \addplot [C0, densely dotted, ultra thin, opacity=0.5, name path=ub] table [x=time, y=ub] {\LowIterData};
            \addplot [C0, densely dotted, opacity=0.2] fill between [of=lb and ub, reverse=true];

            \addplot [C0, densely dashed] table [x=time, y=avg] {\MidIterData};
            \addplot [C0, densely dashed, ultra thin, opacity=0.5, name path=lb] table [x=time, y=lb] {\MidIterData};
            \addplot [C0, densely dashed, ultra thin, opacity=0.5, name path=ub] table [x=time, y=ub] {\MidIterData};
            \addplot [C0, densely dashed, opacity=0.2] fill between [of=lb and ub, reverse=true];

            \addplot [C0] table [x=time, y=avg] {\HighIterData};
            \addplot [C0, ultra thin, opacity=0.5, name path=lb] table [x=time, y=lb] {\HighIterData};
            \addplot [C0, ultra thin, opacity=0.5, name path=ub] table [x=time, y=ub] {\HighIterData};
            \addplot [C0, opacity=0.2] fill between [of=lb and ub, reverse=true];
    \end{groupplot}

    \node[below = 1.2cm of mpcplots c1r1.south] {(a)};
    \node[below = 1.2cm of mpcplots c2r1.south] {(b)};
\end{tikzpicture}
    \caption{\textbf{(a)} Comparison of the evolution of the optimality gap during optimisation (excluding the $N_0$ warm-up iterations) of the MPC tuning problem for myopic method \textit{B} (dashed) and nonmyopic methods \textit{MS-3}, \textit{R-4} (solid). The colour scheme follows from \Cref{fig:numerical:itertime-vs-gap}, and averages and $95\%$ confidence intervals are computed over the 30 repetitions. \textbf{(b)} Evolution  of the reactor temperature $T_\text{R}$, under the MPC policy, during initial (dotted), middle (dashed) and final (solid) iterations of the \textit{MS-3} method, averaged over each of the 30 repetitions of the experiment with $95\%$ confidence intervals. In red, the temperature lower bounds}
    \label{fig:numerical:gaps-and-temp-reactor-mpc-tuning}
\end{figure}

We tested the myopic baseline algorithm \textit{B} \citep{bemporad_global_2020} and the proposed nonmyopic ones as in the previous experiment. In particular, two different nonmyopic approaches were selected as best: the rollout approach with horizon 4,\textit{R-4}, and the multi-step approach with horizon 3, \textit{MS-3}. The results on the convergence of the optimality gap are reported in \Cref{fig:numerical:gaps-and-temp-reactor-mpc-tuning}a, where it is confirmed again that the two nonmyopic methods outperform the myopic one, despite the much higher number of decision variables compared to the benchmarks in \Cref{section:numerical-experiments:benchmarks}. As expected, the multi-step method performs better than the rollout. This result empirically confirms the intuition that, despite a slightly shorter horizon and an even larger number of optimisation variables compared to rollout, branching over multiple possible evolutions of the surrogate model can be beneficial in capturing the real unknown dynamics and thus delivering convergence to a better solution in the same number of iterations.

\Cref{fig:numerical:gaps-and-temp-reactor-mpc-tuning}b shows how the reactor temperature $T_\text{R}$ evolves in time at different optimisation iterations of the \textit{MS-3} method. From this figure, it can be seen that the proposed method is successful in discovering MPC policies that can sustain higher reactor temperatures more consistently. This is advantageous because a higher $T_\text{R}$ will lead to higher reaction rates $k_i$, $i=1,2,3$, leading to a higher conversion of component A to component B. This will in turn maximise the objective \eqref{eq:numerical:mpc-tuning:objective}. At the same time, control policies that induce higher temperatures are less likely to violate the constraint, thus resulting in safer behaviours as the optimisation progresses. This is thanks to these violations being adequately penalised in the objective \eqref{eq:numerical:mpc-tuning:objective}. We note that, although this is an indirect way of addressing constraints, the proposed nonmyopic method can be easily extended to the setting of constrained black-box optimisation \citep{sorourifar_data_2021,airaldi_learning_2023}, a topic that seeks to craft algorithms that are able to systematically handle and learn at the same time unknown constraint functions on the decision variable $\theta \in \Theta$.

\section{Conclusions} \label{section:conclusions}

In this paper, we have demonstrated how GO approaches with deterministic surrogate models can benefit from the incorporation of multi-step lookahead formulations. Building on top of solid theoretic foundations, from dynamic programming to optimal control, we introduced novel nonmyopic acquisition strategies tailored to IDW and RBF surrogate modelling, extending the lookahead paradigm from the Bayesian setting to the deterministic as well. The proposed methods are better suited for handling the complex and ever-present exploration-exploitation dilemma compared to the traditional acquisition functions, and, by acting less greedily, are often able to converge to better solutions in a smaller amount of iterations. Numerical experiments showcased these advantages even in high-dimension, highly nonlinear optimisation benchmarks and problems. As a downside, the proposed methodologies are hindered by the increased computational burden due to the larger and more complex optimisation problems. However, these issues are mitigated by leveraging GPUs and their massive parallelisation capabilities. 

Future work will tackle the issue of constrained global optimisation where the constraint functions are also unknown and need to be represented by IDW or RBF regression models, and investigate how the proposed nonmyopic deterministic strategies can be extended to such scenarios with unknown constraints.

\backmatter

\bmhead*{Author Contributions} 
Conceptualisation: F.A., A.D.; 
Funding Acquisition: A.D., B.D.S.;
Methodology: F.A., A.D.; 
Software: F.A.; 
Visualisation: F.A.; 
Supervision: A.D., B.D.S.;
Writing - original draft preparation: F.A.; 
Writing - review and editing: A.D., B.D.S.

\bmhead*{Funding}
This research is part of a project that has received funding from the European Research Council (ERC) under the European Union's Horizon 2020 research and innovation programme (Grant agreement No. 101018826 - CLariNet).

Funded by the European Union. Views and opinions expressed are however those of the author(s) only and do not necessarily reflect those of the European Union or the European Research Council Executive Agency. Neither the European Union nor the granting authority can be held responsible for them.

\bmhead*{Code Availability}
The source code and simulation results are made available in the following repository: \url{https://github.com/FilippoAiraldi/global-optimization}.

\section*{Declarations}

\bmhead*{Conflict of Interest} 
The authors have no competing interests to declare that are relevant to the content of this article.

\bibliographystyle{apalike}
\bibliography{references-long}

\end{document}